\documentclass[11pt,a4paper]{article}

\usepackage[utf8]{inputenc}
\usepackage[T1]{fontenc}
\usepackage{amsmath,amssymb,amsfonts}
\usepackage{graphicx}
\usepackage{caption}
\usepackage{subcaption}
\usepackage{setspace}
\usepackage{booktabs}
\usepackage{longtable}
\usepackage{array}
\usepackage{multirow}
\usepackage{float}
\usepackage{hyperref}
\usepackage{url}
\usepackage{xcolor}
\usepackage{geometry}
\usepackage[authoryear,round]{natbib}

\newcolumntype{C}[1]{>{\centering\arraybackslash}p{#1}}

\hypersetup{
    colorlinks=true,
    linkcolor=blue,
    filecolor=magenta,      
    urlcolor=cyan,
    citecolor=blue
}

\setcitestyle{numbers,square}

\title{\textbf{GenFAR: A generalized representation of brain structure, derived from 49,246 multi-cohort MRIs via deep learning}}

\author{
Vishnu M. Bashyam\textsuperscript{1*}, Guray Erus\textsuperscript{1}, Junhao Wen\textsuperscript{1}, Pratik Chaudhari\textsuperscript{2},\\
Randa Melhem\textsuperscript{1}, Sindhuja Govindarajan Tirumalai\textsuperscript{1}, Gareth Harman\textsuperscript{1},\\
Yong Fan\textsuperscript{1}, Colin L. Masters\textsuperscript{3}, Paul Maruff\textsuperscript{3}, Sterling C. Johnson\textsuperscript{4},\\
Jurgen Fripp\textsuperscript{5}, Duygu Tosun\textsuperscript{6}, John C. Morris\textsuperscript{7}, Daniel S. Marcus\textsuperscript{8},\\
Pamela LaMontagne\textsuperscript{8}, Tammie Benzinger\textsuperscript{8}, Susan R. Heckbert\textsuperscript{9},\\
Mark Espeland\textsuperscript{10}, Marilyn S. Albert\textsuperscript{11}, Andrew J. Saykin\textsuperscript{12},\\
Paul M. Thompson\textsuperscript{13}, Timothy J. Hohman\textsuperscript{14}, Susan M. Resnick\textsuperscript{15},\\
R. Nick Bryan\textsuperscript{16}, Murat Bilgel\textsuperscript{15}, Yang An\textsuperscript{15}, David A. Wolk\textsuperscript{17},\\
Li Shen\textsuperscript{18}, Haochang Shou\textsuperscript{18}, Ilya M. Nasrallah\textsuperscript{19}, Christos Davatzikos\textsuperscript{1*}\\[12pt]
\parbox{0.7\textwidth}{\raggedright\scriptsize
\textsuperscript{1}Artificial Intelligence in Biomedical Imaging Lab, University of Pennsylvania\\
\textsuperscript{2}Department of Electrical and Systems Engineering, University of Pennsylvania\\
\textsuperscript{3}Florey Institute of Neuroscience and Mental Health, University of Melbourne\\
\textsuperscript{4}Wisconsin Alzheimer's Institute, University of Wisconsin School of Medicine and Public Health\\
\textsuperscript{5}CSIRO Health and Biosecurity, Australian e-Health Research Centre CSIRO\\
\textsuperscript{6}Department of Radiology and Biomedical Imaging, University of California, San Francisco\\
\textsuperscript{7}Department of Neurology, Washington University in St. Louis\\
\textsuperscript{8}Department of Radiology, Washington University in St. Louis\\
\textsuperscript{9}Cardiovascular Health Research Unit and Department of Epidemiology, University of Washington\\
\textsuperscript{10}Department of Biostatistics and Data Science, Wake Forest School of Medicine\\
\textsuperscript{11}Department of Neurology, Johns Hopkins University School of Medicine\\
\textsuperscript{12}Indiana Alzheimer's Disease Research Center, Indiana University\\
\textsuperscript{13}Imaging Genetics Center, Mark and Mary Stevens Neuroimaging and Informatics Institute, Keck School of Medicine, University of Southern California\\
\textsuperscript{14}Vanderbilt Memory and Alzheimer's Center, Vanderbilt University School of Medicine\\
\textsuperscript{15}Laboratory of Behavioral Neuroscience, National Institute on Aging\\
\textsuperscript{16}Department of Diagnostic Medicine, University of Texas at Austin\\
\textsuperscript{17}Department of Neurology, University of Pennsylvania\\
\textsuperscript{18}Department of Biostatistics, Epidemiology and Informatics, University of Pennsylvania\\
\textsuperscript{19}Department of Radiology, University of Pennsylvania}
}

\date{}

\begin{document}

\maketitle

% Abstract (single-spaced)
\singlespacing
\begin{abstract}
Deep learning models for neuroimaging have largely been developed for individual tasks, limiting knowledge transfer across applications. Here we introduce GenFAR, a modular deep learning framework that learns general, clinically informed features from brain MRIs. We trained this modular architecture on 49,246 individuals across 11 cohorts, using 17 diverse classification and regression tasks spanning cognition, clinical, diagnosis, demographics, and biomarkers. This yields aggregated, focused feature sets that capture rich, clinically- and biologically-relevant brain representations. We developed a sequential learning approach where tasks progressively build on previously learned representations. Through an analysis of 5,000 task sequences, we identified an optimal sequence length of six tasks and introduced a Donor Score metric to quantify each task's contribution to downstream performance. This analysis revealed five consistently strong donor tasks (Age, AD/MCI, MMSE, Hypertension, Hyperlipidemia) that formed the base of our sequential model. We demonstrated the utility of our learned representation, in various tasks beyond those included in the training set, to serve as the foundation for specialized secondary predictors. We further showed that using the learned feature representation can substantially increase the sample efficiency of secondary deep learning training tasks and models, as well as improve their accuracy.
\end{abstract}

% Main sections (double-spaced)
\doublespacing
\section*{Introduction}

Over the past decade, deep learning has transformed medical imaging research, leading to significant advancements across various applications, including segmentation, registration, classification, regression, clustering, synthesis, and more \cite{avbersek2022,yao2020}.
With the aid of large datasets and methodological innovations, deep learning has emerged as a critical tool for enabling the clinical translation of computational applications in medical imaging.
However, the current landscape of deep learning models in medical imaging is marked by specialization, with most models tailored for individual use cases and tasks \cite{avbersek2022}.
This specialization, while yielding impressive single-task performance, limits cross-task knowledge transfer, requires substantial engineering effort for each new endpoint, and constrains generalization to diverse clinical populations and acquisition settings.

In recent years, the deep learning community, particularly in the fields of natural language processing and computer vision, has shifted towards larger, more general, foundational models \cite{bommasani2021,ramesh2021,brown2020,openai2023,rombach2022,scao2022}.
These models have demonstrated unprecedented performance and adaptability by leveraging massive datasets, substantial model capacity, and extensive optimization budgets, surpassing the typical diminishing returns associated with increasing scale \cite{bommasani2021}.
This shift has given rise to a new paradigm in deep learning, where novel and specialized tasks are built upon the general feature representations learned by foundational models.

Medical imaging stands to benefit substantially from this paradigm \cite{moor2023}.
Relative to natural images, factors such as small sample sizes, high dimensionality, data homogeneity, and shared imaging features among different pathologies make cross-task and cross-dataset knowledge transfer particularly impactful.
The transition towards foundational models with highly general feature representations directly addresses the current fragmentation of deep learning model development in medical imaging.
Specialized secondary models can benefit from indirect access to larger, diverse datasets and cross-task knowledge transfer, via the sample efficiency provided by learned feature representations \cite{bengio2013}.
These rich image representations, which stem from sizeable and diverse data sources, are expected to improve the performance and generalizability of the models, ultimately benefiting a wide range of medical imaging applications.

Prior work leveraging cross-task knowledge transfer in medical imaging has often focused on the benefits of integrating object detection and segmentation tasks to improve a single classification task.
Multiple publications have demonstrated the benefits of this approach across a variety of imaging modalities and anatomical regions \cite{cao2018,cheng2022,chowdary2022,gao2020,kuang2021,chelaramani2021}.
Other works have investigated the complementary benefits of jointly learning multiple closely related tasks, to improve performance on a target task \cite{li2019,tang2020,zhang2022,he2022}.
While these works offer a promising view into what is possible with cross-task knowledge transfer, they are limited to a small number of closely related tasks and, critically, do not evaluate the performance of new, unseen tasks based on prior learning.

More recently, foundational-model directions have begun to emerge within neuroimaging, with several recent efforts being highly relevant to our work.
Tak et al. introduced a generalized brain MRI foundation model, called BrainIAC, using contrastive self-supervision across 48,519 scans from diverse datasets, reporting strong performance, particularly in low-data regimes, on downstream targets such as age prediction, glioma IDH mutation status, and survival \cite{tak2024brainiac}.
Similarly, Kaczmarek et al. developed a SimCLR-based contrastive learning foundation model for 3D brain structural MRI, pre-trained on 18,759 patients (44,958 scans) from 11 publicly available datasets spanning diverse neurological diseases including Alzheimer's Disease, Parkinson's Disease, and stroke \cite{kaczmarek2025simclr}.
Their model demonstrated superior performance compared to Masked Autoencoders and supervised baselines across multiple downstream tasks, achieving comparable performance to supervised methods when fine-tuning on only 20\% of the data.
Wood et al. (BrainAge) trained modality-specific age predictors on up to 18,890 clinically acquired MR examinations and showed that these models can serve as foundation models whose features transfer effectively (with fine-tuning) to new sequences, orientations, and even modest sample sizes \cite{wood2024brainage}.
Beyond brain-specific models, Wang et al. (Triad) trained a single 3D encoder over 131,170 MRIs spanning multiple organs and modalities and attached task-specific heads to address a broad battery of segmentation, classification, and registration tasks, demonstrating state-of-the-art results within-domain and non-trivial zero-shot transfer \cite{wang2025triad}.
Complementing these, Deng et al. (SAM-Brain3D) presented a brain-focused foundation model based on a 3D SAM architecture with an adapter for downstream brain disease analysis, emphasizing segmentation at scale and extension to classification tasks via lightweight adapters \cite{deng2025sambrain3d}.
Related segmentation-focused models (BrainFounder \cite{cox2024brainfounder}, LaMiM \cite{chen2024lamim}) and robust segmentation frameworks with strong cross-contrast generalization (SynthSeg \cite{billot2023synthseg}) further attest to the feasibility of brain-imaging foundation models.

These developments reveal diverse approaches to learning generalizable neuroimaging representations, each capturing different types of priors.
BrainIAC, Triad, and the SimCLR-based model by Kaczmarek et al. learn powerful priors using large-scale self-supervision and adapt with task-specific heads.
Segmentation-centric foundation models (BrainFounder, LaMiM, SynthSeg and SAM-Brain3D) also learn priors but through supervised segmentation tasks that capture structural regularities of brain anatomy.
These approaches have demonstrated impressive results, however, they are not designed to explicitly encode clinically informed, task-specific supervision from a wide spectrum of neuroimaging phenotypes nor to study how learning across such phenotypes can be structured to maximize transfer within the brain domain.
BrainAge \cite{wood2024brainage} provides an encouraging proof-of-concept, showing that a single clinically motivated endpoint (age prediction) can produce broadly useful features with strong transfer capabilities even in limited data settings.
This raises the natural question of whether jointly learning from multiple clinical endpoints might capture richer, more generalizable representations than any single task alone.
Lastly, the meta-matching framework by He et al. explores multi-phenotype learning from a different angle, training separate predictors and reusing their outputs as features for new targets, yielding notable sample-efficiency on external cohorts \cite{he2022}.
While impressive, it operates on final predictions rather than intermediate representations and does not support explicit knowledge sharing during the feature-learning stage.

In this work, we set out to learn clinically informed, general neuroimaging representations from 3D T1-weighted MRIs by supervising on a broad set of phenotypes and by structuring how tasks learn from one another.
We refer to our approach as GenFAR (Generalized FeAture Representations).
We explored two complementary frameworks for this approach.
The first, an independent framework, employs a modular model comprised of multiple task-specific prediction channels trained in parallel without inter-task knowledge sharing.
These channels learn focused features that, when aggregated, form a general representation of the brain (Figure 1B-C).
These independently learned features can transfer well to new tasks and markedly increase sample-efficiency, often outperforming directly trained models in low-data regimes.
However, this independent framework does not allow information to flow between tasks during representation learning, which means weaker tasks with limited data or tasks with inherently low separability on T1-MRI cannot benefit from the structure learned by stronger tasks.

To address this limitation, we also developed a sequential learning framework that enables tasks to learn incrementally from previously learned representations.
In this framework, tasks are arranged in an ordered sequence, where each task receives both the 3D brain scan and the feature representations from all preceding tasks, allowing later tasks to bootstrap from stronger, upstream tasks rather than re-learning from scratch.
This design preserves the modularity of task-specific channels while introducing a controlled mechanism for knowledge transfer across tasks during training.
Our sequential learning approach draws heavily from the Model Zoo framework \cite{ramesh2021model}, which demonstrated the effectiveness of growing an ensemble of models trained on different task combinations to mitigate task competition in continual learning.
The sequential learning framework requires careful consideration of sequence length to balance transfer against noise accumulation and overfitting, as well as strategic task ordering to maximize downstream benefit.
Through extensive empirical analysis of 5,000 random sequences spanning a range of lengths, we identified an optimal sequence length of six and introduced a Donor Score to quantify a task's marginal benefit to others.
This analysis revealed a small set of consistently strong donors, i.e. Age, AD/MCI (Alzheimer's Disease/Mild Cognitive Impairment), MMSE (Mini-Mental State Examination), Hypertension, Hyperlipidemia, which we use to construct a robust base sequence.

We evaluated GenFAR under both learning frameworks across 17 clinically relevant tasks encompassing cognition, neurologic diagnosis, demographics, various risk factors, and cerebrospinal fluid (CSF) biomarkers.
The models were trained on a large, heterogeneous dataset of 49,246 3D T1-weighted brain MRIs drawn from 11 studies and multiple scanner vendors and protocols, with minimal preprocessing to preserve full spatial context.
We assessed generalization using a leave-task-out scheme that holds out each task in turn and trains secondary predictors solely on the features learned from the remaining tasks.
We further examined sample-efficiency by restricting the downstream training set size and evaluated study-to-study generalization on a separate external cohort.
Across these settings, we found that the sequential framework consistently improved upon the independent framework and direct training, with especially pronounced gains in low-data regimes and for classification endpoints.
The independent channels remain a strong baseline and a modular component of the overall GenFAR design.
Taken together, GenFAR contributes a clinically anchored, brain-specific foundation for predictive modeling that complements recent self-supervised and segmentation-centric brain models by learning generalizable features from many supervised neuroimaging endpoints, providing a principled mechanism for structured knowledge transfer via sequential learning, and yielding practical gains in accuracy and sample-efficiency on unseen tasks, within and across studies.

\begin{figure}[htbp]
\centering
\includegraphics[width=0.95\textwidth]{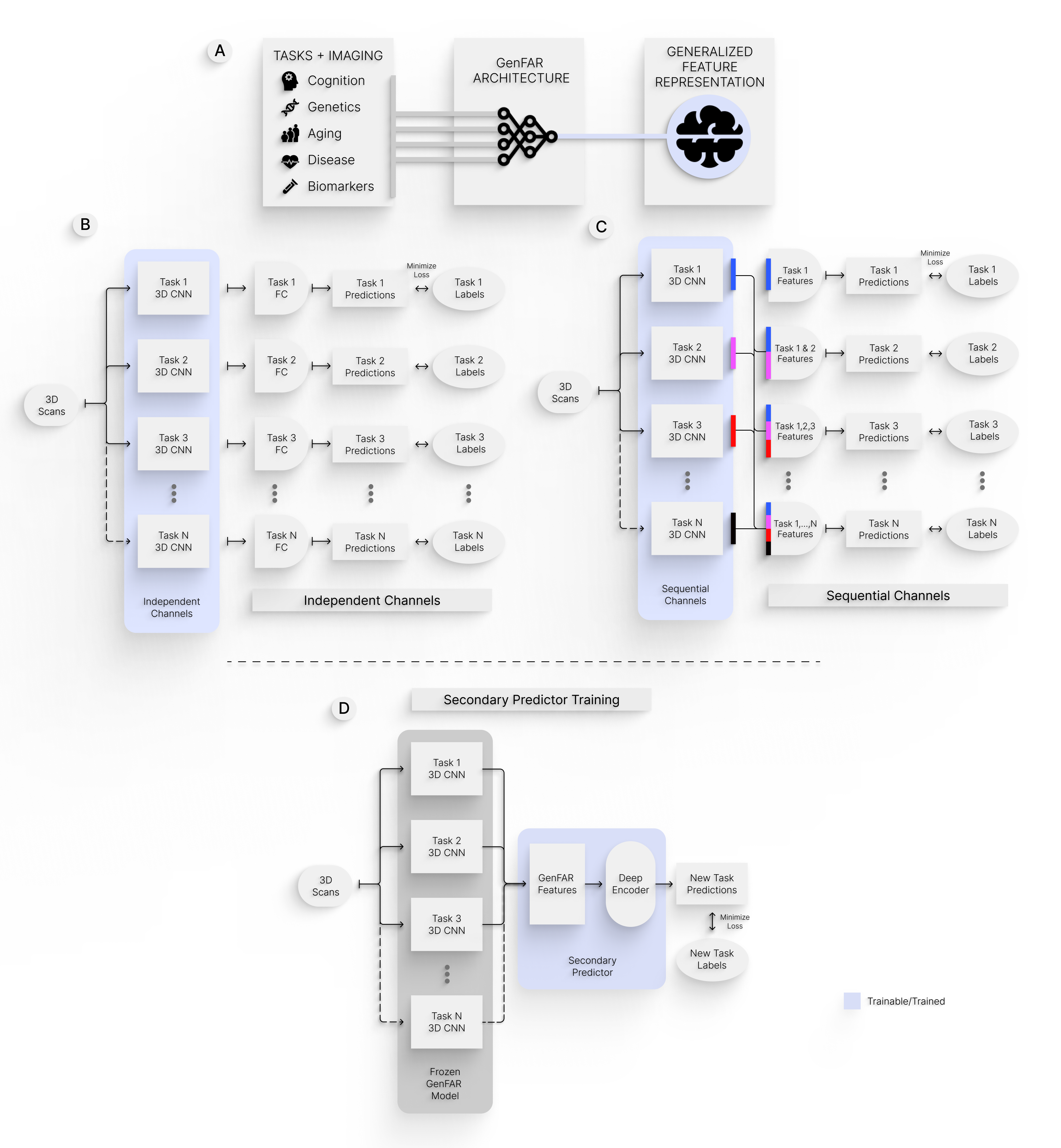}
\caption{\textbf{Overview of the GenFAR model architecture and training frameworks.} \textbf{(A)} GenFAR leverages 3D T1-weighted brain MRI scans paired with diverse clinical labels spanning cognition, genetics, aging, disease, and biomarkers to learn generalizable feature representations. The modular architecture leverages heterogeneous neuroimaging tasks to produce a unified, clinically-informed feature space. \textbf{(B)} Independent learning framework showing 17 parallel task-specific prediction channels operating independently. Each channel consists of a 3D Squeeze-and-Excitation (SE) ResNet followed by fully connected (FC) layers optimized for their respective tasks. Each channel processes the 3D brain scan and learns a 512-dimensional feature representation without inter-task knowledge sharing. Channels are optimized end-to-end using task-appropriate loss functions (binary cross-entropy for classification, mean squared error for regression). \textbf{(C)} Sequential learning framework where tasks are arranged in an optimized sequence. Later tasks receive both the 3D brain scan and concatenated feature representations from all preceding tasks (shown as colored bars), enabling progressive knowledge transfer. The first task operates identically to the independent framework, while subsequent tasks incrementally build upon previously learned representations. \textbf{(D)} Secondary predictor training for novel tasks. After primary training, convolutional weights of all task-specific channels are frozen to create a fixed GenFAR feature extractor. For new prediction tasks, 3D scans pass through the frozen GenFAR model to generate comprehensive feature representations (8,704 features for independent framework; 3,200 features for sequential framework). These features serve as input to a lightweight trainable deep encoder optimized for the novel task. This architecture enables efficient transfer learning, improving sample efficiency and performance on new tasks by leveraging representations learned from large-scale, diverse training.}
\end{figure}

A key contribution of this work lies in development of highly generalizable neuroimaging feature representations. To maximize their utility for the research community, the final model, along with a self-contained GPU and CPU inference pipelines, are publicly released on GitHub (\url{https://github.com/vishnubashyam/GenFAR_Main}) and released under an Open Responsible AI License (OpenRAIL). This accessible pipeline allows researchers to easily extract these optimized, generalized features from their own raw neuroimaging scans, to be used in subsequent data analysis steps.
We also provide a web portal via NiChart (\url{https://neuroimagingchart.com/portal}) to encourage direct use of GenFAR on the cloud without installation or the need for computing resources. The web portal, hosted in the Amazon Web Services cloud, allows users to submit their raw T1 images by a simple drag and drop interface and to get back the extracted GenFAR feature set.

\section*{Results}

\subsection*{Task-specific prediction channels learn specialized representations that are predictive of their target task}

We first evaluated the performance of each prediction channel on its corresponding target task to establish a performance baseline. This was done for two distinct learning frameworks: (1) an independent framework, where each of the 17 prediction channels was trained in parallel without knowledge sharing, and (2) a sequential framework, designed to allow tasks to build upon previously learned feature representations.

Using independent prediction channels, each prediction channel was trained and evaluated directly on its corresponding task using 3D T1-weighted brain MRIs. We observed that certain tasks exhibited substantially higher predictive performance than other tasks. This variability could be due to a number of factors, such as the number of training samples available, the difficulty of the task, and the extent to which the input data is informative about the task. Many of the prediction channels are highly performant on their respective tasks; however, our experiments revealed that even in cases where the task-specific performance of certain features may be relatively low, the learned feature representations may be useful in other tasks. The observed results are likely because models trained on distinct clinical tasks (e.g., identifying smoking and diabetes) are not expected to yield fully separable neuroimaging features, but rather to capture the inherent, shared neurobiological overlap between subjects.

Additionally, we developed a sequential learning framework where tasks learn incrementally from previously learned representations. In this setup, each task in a sequence receives both the 3D brain scan and the feature representations from all preceding tasks, allowing later tasks to bootstrap from stronger, earlier tasks rather than learning from scratch. To determine optimal configurations for sequential learning, we conducted extensive experiments with 5,000 random task sequences of varying lengths (3-16 tasks). Initial attempts at optimization-based approaches (Thompson sampling and evolutionary algorithms) resulted in severe overfitting to the validation set due to the repeated evaluations during optimization. Instead, we examined aggregate patterns across random sequences to identify optimal parameters.

Our analysis revealed that sequences of length 6 demonstrated statistically significant superior performance compared to other lengths (p < 0.05 after FDR correction), proving to be statistically superior to both shorter sequences (Length 1 and 3) and longer sequences (Lengths 9–16). Longer sequences, particularly those of length 9 and above, exhibited extreme overfitting with strong validation performance but poor test generalization (\textbf{Figure 2}). This optimal length balances the benefits of knowledge transfer against the accumulation of noise and overfitting that occurs in longer sequences.

\begin{figure}[H]
\centering
\includegraphics[width=0.95\textwidth]{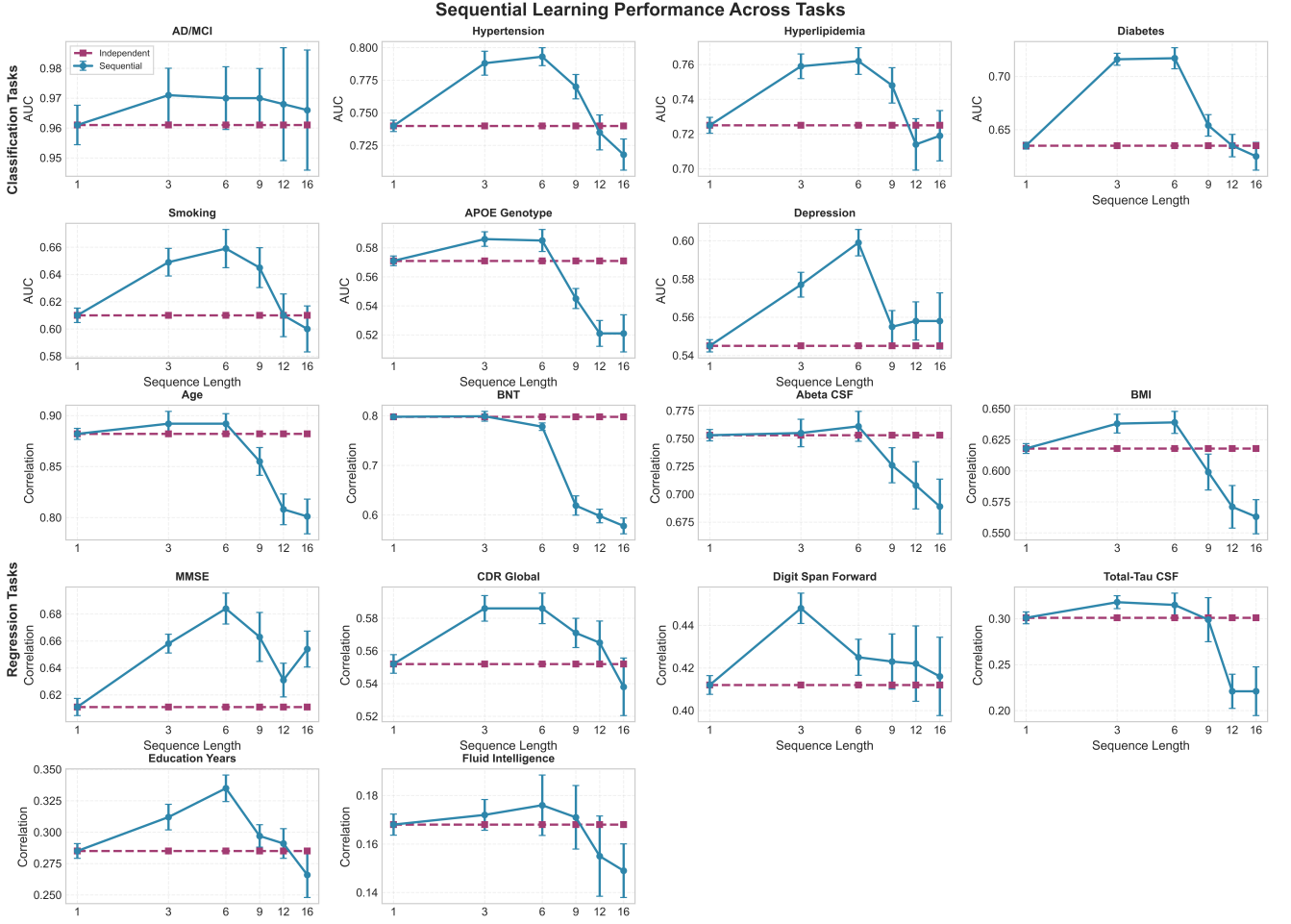}
\caption{\textbf{Sequential learning performance across varying sequence lengths.} Performance improvements relative to baseline were evaluated across 5,000 randomly generated task sequences of lengths 1, 3, 6, 9, 12, and 16. Each point represents the mean performance improvement for all sequences of that length. Bars indicate standard deviation. Sequences exceeding length 6 show progressive performance degradation, with lengths 12 and 16 exhibiting sharp declines due to overfitting. The validation set performance (not shown) remains high for longer sequences, confirming that the performance drop reflects overfitting rather than training failure. AD, Alzheimer's Disease; MCI, Mild Cognitive Impairment; MMSE, Mini-Mental State Examination; CDR®, Clinical Dementia Rating®; BNT, Boston Naming Test; CSF, cerebrospinal fluid; BMI, body mass index; APOE, apolipoprotein E.}
\end{figure}

The sequential learning channels, using the optimized sequence length of 6, compared to the independent channels, are summarized in \textbf{Table 1}. The sequential framework generally enhanced prediction performance across most tasks compared to the independent framework. For classification tasks, this led to consistent improvements, with the most substantial gains in AUC for Diabetes (0.635 to 0.717) and Hypertension (0.740 to 0.781). A similar trend was observed for regression tasks, where sequential learning improved correlation for 9 out of 10 tasks, most notably for MMSE (0.611 to 0.681). While not all tasks benefited (BNT showed a minor performance decrease) the overall results demonstrate that the sequential framework effectively leverages shared information to boost performance.

\newpage
\singlespacing
\begin{longtable}{lcccccc}
\toprule
\multicolumn{7}{c}{\textbf{Classification Tasks}} \\
\midrule
Task & Task Type & \multicolumn{2}{c}{AUC} & \multicolumn{2}{c}{Accuracy} & Num. Samples \\
\cmidrule(lr){3-4} \cmidrule(lr){5-6}
& & Individual & Sequential & Individual & Sequential & \\
\midrule
AD/MCI & Classification & 0.961 & 0.97 & 0.952 & 0.950 & 5611 \\
Hypertension & Classification & 0.74 & 0.781 & 0.737 & 0.786 & 22711 \\
Hyperlipidemia & Classification & 0.725 & 0.770 & 0.733 & 0.752 & 17466 \\
Smoking & Classification & 0.61 & 0.661 & 0.607 & 0.634 & 45002 \\
APOE Genotype & Classification & 0.571 & 0.585 & 0.566 & 0.573 & 40942 \\
Diabetes & Classification & 0.635 & 0.717 & 0.552 & 0.648 & 4106 \\
Depression & Classification & 0.545 & 0.587 & 0.514 & 0.570 & 4506 \\
\midrule
\multicolumn{7}{c}{\textbf{Regression Tasks}} \\
\midrule
Task & Task Type & \multicolumn{2}{c}{Correlation} & \multicolumn{2}{c}{MAE} & Num. Samples \\
\cmidrule(lr){3-4} \cmidrule(lr){5-6}
& & Individual & Sequential & Individual & Sequential & \\
\midrule
Age & Regression & 0.882 & 0.892 & 3.72 & 3.7 & 49246 \\
Boston naming test & Regression & 0.798 & 0.778 & 4.02 & 4.1 & 4565 \\
Abeta CSF& Regression & 0.753 & 0.761 & 53.98 & 52.23 & 1691 \\
Body mass index & Regression & 0.618 & 0.639 & 2.63 & 2.54 & 46840 \\
MMSE & Regression & 0.611 & 0.681 & 1.97 & 1.66 & 6163 \\
CDR Global & Regression & 0.552 & 0.596 & 0.21 & 0.179 & 4674 \\
Digit Span Forward & Regression & 0.412 & 0.425 & 1.28 & 1.31 & 31014 \\
Total-Tau CSF & Regression & 0.301 & 0.315 & 38.45 & 38.21 & 1673 \\
Education Years & Regression & 0.285 & 0.335 & 2.32 & 2.08 & 6963 \\
Fluid Intelligence & Regression & 0.168 & 0.191 & 1.63 & 1.49 & 36435 \\
\bottomrule
\\[12pt] % Add space before caption
\nopagebreak % Prevent page break before caption
\multicolumn{7}{p{0.95\textwidth}}{\small \textbf{Table 1: Task-Specific Prediction Channel Performance.} The test set performance for each prediction task is shown, comparing models trained on individual channels versus sequential channels (sequence length of 6). For classification tasks, area under the curve (AUC) and accuracy are used to quantify performance. For regression tasks, the Pearson correlation coefficient between the predicted and actual values and mean absolute error (MAE) are used to quantify performance. The inclusion of sequential data generally enhances prediction performance across most tasks, particularly for classification. However, certain tasks may exhibit varied levels of improvement, reflecting the differential impact of sequential information on task difficulty. AD, Alzheimer's Disease; MCI, Mild Cognitive Impairment; MMSE, mini-mental state examination; CDR®, clinical dementia rating scale; CSF, cerebrospinal fluid.} \\
\end{longtable}
\doublespacing

\subsection*{Task ordering reveals that select tasks disproportionately benefit downstream performance}

To systematically evaluate how individual tasks contribute to sequential learning, we developed the Donor Score metric (\textbf{Figure 3A}).
This metric quantifies the marginal benefit a task provides when included in a learning sequence compared to sequences without it.

For a given receiver task, the Donor Score partitions all learning sequences into two groups: those containing a specific donor task anywhere in the sequence preceding the final task, and those that do not.
The score is calculated as the difference between the average percent performance improvement for sequences with the donor task versus those without it.
A positive score indicates beneficial or synergistic relationships, while negative scores suggest antagonistic interactions where the donor task is detrimental to the receiver's performance.
This approach allows us to quantify the contribution of each task as a feature donor across all possible receiver tasks, providing insights into the underlying structure of knowledge transfer in our multi-task framework.

Analysis of pairwise Donor Scores across all tasks revealed distinct patterns of task relationships (\textbf{Figure 3B}).
Certain tasks emerged as consistently strong positive donors across multiple receiver tasks: Age (average Donor Score: 0.12), AD/MCI (0.10), MMSE (0.09), Hypertension (0.07), and Hyperlipidemia (0.06).
These tasks rarely produced negative donations and provided broad benefits across diverse prediction targets.

Conversely, certain tasks exhibited predominantly negative Donor Scores (\textbf{Figure 3C \& 3D}), most notably Total-Tau CSF and Smoking.
For Total-Tau CSF, this pattern likely reflects limited training data leading to noise and overfitting, while Smoking's negative contributions may stem from weak structural correlates on T1-MRI that introduce uninformative variance rather than meaningful generalizable features.

Based on these findings, we established a fixed base sequence using the five strongest donor tasks (Age, AD/MCI, MMSE, Hypertension, Hyperlipidemia) for our final sequential model.
Each of the remaining 10 non-base tasks (excluding Total-Tau CSF and Smoking due to their strongly negative Donor Scores) was then inserted as the sixth task in this sequence, allowing it to benefit from the strong base features while preventing the accumulation of noise.

\begin{figure}[htbp]
\centering
\includegraphics[width=\textwidth]{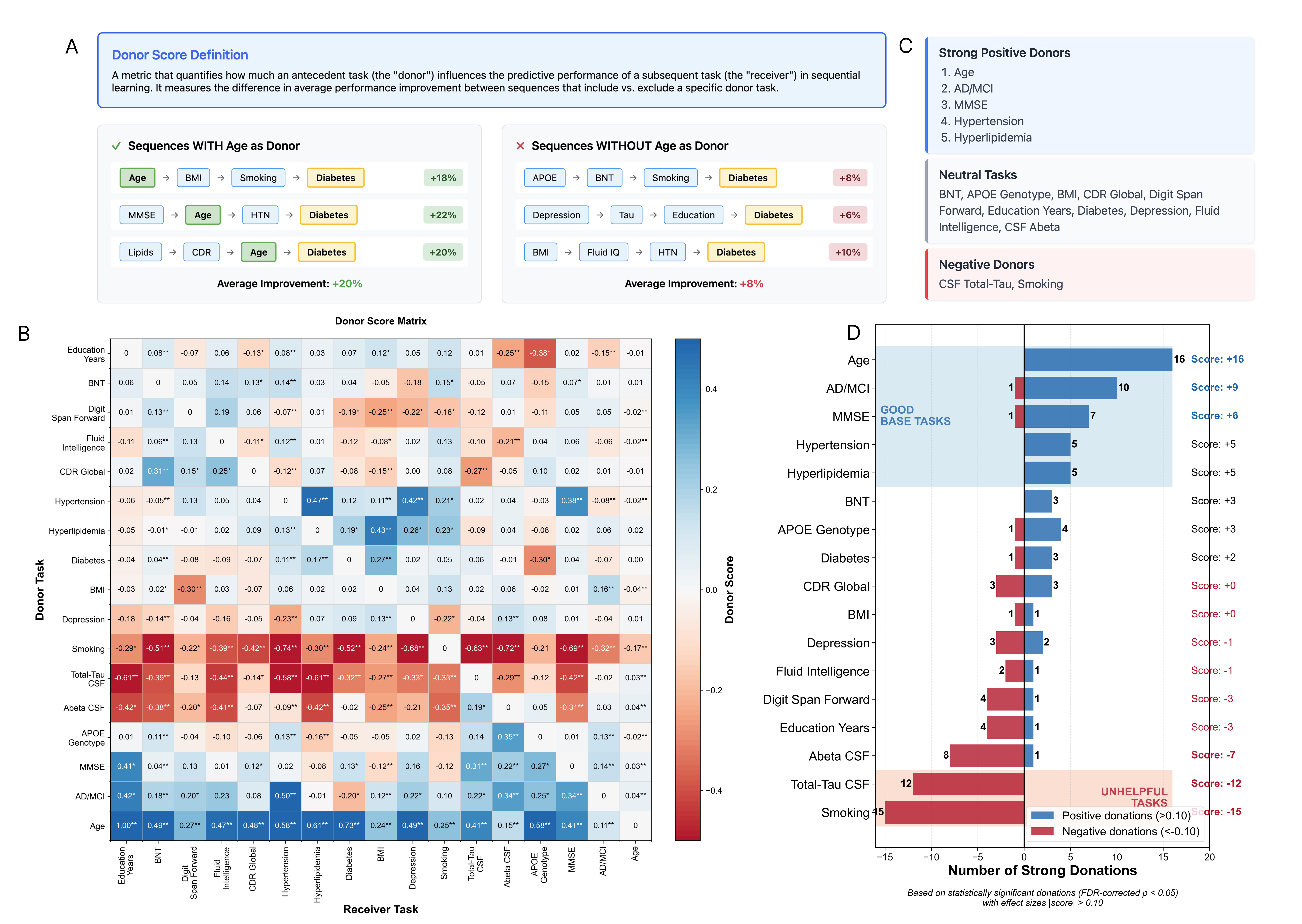}
\caption{\textbf{Donor Score analysis reveals task relationships in sequential learning.} \textbf{(A)} Schematic illustration of the Donor Score calculation using Age as the example donor task and Diabetes as the example receiver task. \textbf{(B)} Heatmap of pairwise Donor Scores quantifying the influence of donor tasks on subsequent receiver tasks. Each cell represents the difference in average performance improvement between sequences containing versus excluding the donor task. Warm colors indicate positive contributions; cool colors indicate negative effects. Asterisks denote statistical significance after FDR correction (*p < 0.05, **p < 0.001). \textbf{(C)} Task categorization based on Donor Score patterns. Strong positive donors (Donor Score > 0.1) consistently benefit downstream tasks; neutral tasks show mixed effects; negative donors (Donor Score < -0.1) impair downstream performance, often due to limited training data. \textbf{(D)} Distribution of strong donations across tasks. Bar length indicates the number of receiver tasks for which each donor provides statistically significant benefit (blue) or detriment (red). Tasks are ordered by net contribution, highlighting Age, AD/MCI, MMSE, Hypertension, and Hyperlipidemia as universally beneficial donors that form the foundation of the final sequential learning framework.}
\end{figure}

\subsection*{Jointly, the task-specific prediction channels learn representations that are highly generalizable to new tasks}

To assess the applicability and generalizability of the learned features to novel tasks, we employed Leave-Task-Out Cross-Validation (LTO-CV). This validation process isolated one task as a test set while utilizing the remaining tasks as the training set for task-specific prediction channels. The predictive performance of the derived features was then evaluated on the held-out task. The performance on the held-out task does not benefit from direct training and is restricted to the generalizability of features learned in other tasks.

Since some scans may have multiple task labels associated with them, a specific sampling strategy was used to prevent data leakage to the held-out task (See Methods - Leave-Task-Out Cross-Validation section). Given this sampling, the number of samples used to evaluate the target task may be fewer than the number used in the task-specific performance evaluation reported above. For both independent and sequential approaches, we compared performance against directly trained baselines using identical data splits (\textbf{Table 2}).

\singlespacing
\begin{longtable}{lccccccc}
\toprule
\multicolumn{8}{c}{\textbf{Classification Tasks}} \\
\midrule
Task & \multicolumn{3}{c}{AUC} & \multicolumn{3}{c}{Accuracy} & Samples \\
\cmidrule(lr){2-4} \cmidrule(lr){5-7}
& Direct & Independent & Sequential & Direct & Independent & Sequential & \\
\midrule
AD/MCI & 0.961 & 0.967 & 0.97 & 0.952 & 0.947 & 0.952 & 5611 \\
Hypertension & 0.64 & 0.762 & 0.793 & 0.579 & 0.752 & 0.784 & 8000 \\
Hyperlipidemia & 0.687 & 0.741 & 0.762 & 0.637 & 0.737 & 0.758 & 8000 \\
Diabetes & 0.635 & 0.733 & 0.747 & 0.552 & 0.729 & 0.748 & 4106 \\
Smoking & 0.549 & 0.622 & 0.659 & 0.511 & 0.621 & 0.645 & 8000 \\
APOE Genotype & 0.509 & 0.582 & 0.588 & 0.506 & 0.575 & 0.581 & 8000 \\
Depression & 0.545 & 0.562 & 0.599 & 0.514 & 0.558 & 0.595 & 4506 \\
\midrule
\multicolumn{8}{c}{\textbf{Regression Tasks}} \\
\midrule
Task & \multicolumn{3}{c}{Correlation} & \multicolumn{3}{c}{MAE} & Samples \\
\cmidrule(lr){2-4} \cmidrule(lr){5-7}
& Direct & Independent & Sequential & Direct & Independent & Sequential & \\
\midrule
Age & 0.835 & 0.793 & 0.812 & 4.12 & 4.62 & 4.28 & 8000 \\
Abeta CSF & 0.753 & 0.721 & 0.741 & 53.98 & 54.12 & 53.84 & 1691 \\
MMSE & 0.611 & 0.652 & 0.694 & 1.97 & 1.85 & 1.53 & 6163 \\
BMI & 0.528 & 0.561 & 0.579 & 3 & 2.96 & 2.88 & 8000 \\
CDR Global & 0.552 & 0.561 & 0.586 & 0.21 & 0.2 & 0.185 & 4674 \\
BNT & 0.798 & 0.558 & 0.628 & 4.02 & 4.48 & 4.18 & 4565 \\
Digit Span Forward & 0.406 & 0.415 & 0.413 & 1.27 & 1.25 & 1.31 & 8000 \\
Education Years & 0.285 & 0.289 & 0.315 & 2.32 & 2.32 & 2.13 & 6963 \\
Total-Tau CSF & 0.301 & 0.253 & 0.274 & 38.44 & 38.98 & 38.73 & 1673 \\
Fluid Intelligence & 0.119 & 0.161 & 0.176 & 1.64 & 1.61 & 1.57 & 8000 \\
\bottomrule
\\[12pt] % Add space before caption
\nopagebreak % Prevent page break before caption
\multicolumn{8}{p{0.95\textwidth}}{\small \textbf{Table 2: Evaluating the generalizability of learned features to novel tasks using Leave-Task-Out Cross-Validation (LTO-CV).} Performance comparison of three approaches: Direct training on 3D MRI data, Independent GenFAR features, and Sequential GenFAR features. LTO-CV was performed by holding out one task, training all other task-specific prediction channels, then evaluating the predictive performance of the learned features on the held-out task. Direct columns show performance when the network is trained directly on the target prediction task using the same samples reserved for the held-out task in the LTO experiment. Independent columns show performance using features learned from other tasks without sequential knowledge transfer. Sequential columns show performance with sequential learning where tasks build upon previously learned representations. For classification tasks, area under the curve (AUC) and accuracy quantify performance. For regression tasks, Pearson correlation coefficient and mean absolute error (MAE) quantify performance.} \\
\end{longtable}
\doublespacing

Among classification tasks (\textbf{Table 2}, top section), the sequential LTO experiments outperformed direct training in all 7 tasks, with particularly substantial improvements ($\geq$10\% accuracy) in Hypertension, Hyperlipidemia, Smoking, and Diabetes. For regression tasks (\textbf{Table 2}, bottom section), sequential LTO outperformed direct training in 6 of 10 tasks, with notable improvements in MMSE (correlation: 0.694 vs. 0.611), BMI (0.579 vs. 0.528), and Education Years (0.315 vs. 0.285). Of the four remaining tasks where direct training performed better, three showed relatively modest differences: Age (correlation: 0.812 vs. 0.835), Abeta CSF (0.741 vs. 0.753), and Total-Tau CSF (0.274 vs. 0.301), with correlation coefficient differences of less than 0.03. Only the Boston Naming Test (BNT) showed a substantial performance gap where sequential LTO notably underperformed direct training (correlation: 0.628 vs. 0.798), representing the single task with a meaningful performance deficit ($\geq$0.1 correlation coefficient difference) when using learned features.

Overall, in 16 of 17 tasks evaluated, models using sequential learned features achieved comparable or superior performance to models trained with full access to the original imaging data. Additionally, the sequential learning framework demonstrated consistent improvements over the independent channels across nearly all tasks.

\subsection*{GenFAR substantially increases sample efficiency in secondary prediction tasks over directly trained networks}

While the large amount of training data used in previous sections is essential for learning generalized feature representations, they are atypical of the sample sizes commonly encountered in neuroimaging studies\cite{avbersek2022,yao2020}. Therefore, we examined the sample efficiency, defined as the ability to learn a predictive signal from limited data, of our method. We evaluated the sample efficiency of secondary predictors using the same framework as the leave-task-out cross-validation scheme while limiting the number of training samples that are available in the secondary step. In this way, we attempted to determine the comparative benefit of using GenFAR features on a new task with limited training data.

We evaluated the performance of independent versus sequential training strategies across a limited range of sample sizes, beginning with 4,000 subjects and systematically decreasing down to 100. The sequential learning framework demonstrated superior performance across limited sample sizes compared to both independent GenFAR features and direct training across all sample size ranges (\textbf{Figure 4}).

For classification tasks, GenFAR features consistently achieved the strongest performance across all sample ranges, maintaining predictive performance for longer as sample size decreased. This result was particularly pronounced in AD/MCI, Hyperlipidemia, Hypertension, and Diabetes tasks. The average AUC gain of sequential over direct training was relatively comparable across sample size ranges (100-1000, 1000-2000, 2000-4000). Relative to the independent features, the sequential framework showed slight average AUC improvements across all ranges For regression tasks, sequential learning also exceeded both alternatives across all ranges, with the largest correlation gains appearing at the smallest data scales. The average improvement over direct training was highest in 100-1000 sample range and remained sizable in the 1000-4000 sample range.

Notably, the four tasks where direct training outperformed sequential learning at full sample size (Age, Abeta CSF, Total-Tau CSF, and BNT) all showed a crossover point where sequential features provided higher prediction performance as training data decreased. Sequential learning surpassed direct training for Age below 2000 samples, for Abeta CSF at 1000 samples, and for both BNT and Total-Tau CSF at 500 samples.

\begin{figure}[htbp]
\centering
\includegraphics[width=0.95\textwidth]{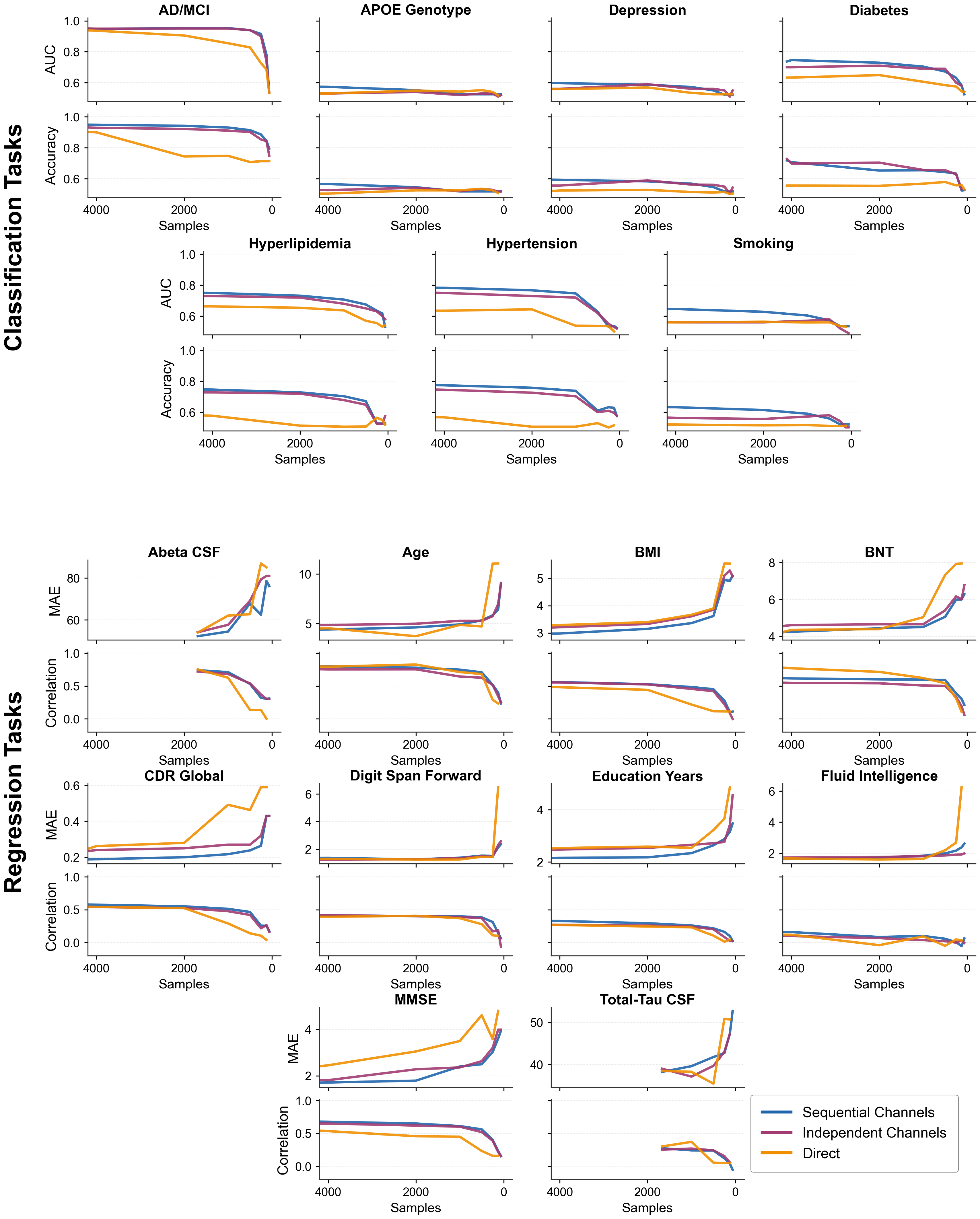}
\caption{\textbf{Sample efficiency using GenFAR features in a secondary predictor versus direct training.} Performance comparison of three approaches as training data is progressively restricted to simulate limited-data scenarios typical of neuroimaging studies. Classification task performance (Upper panel) measured by AUC and accuracy as a function of decreasing training sample size. Regression task performance (Lower panel) measured by correlation coefficient and MAE as a function of decreasing training sample size. The performance advantage of sequential and independent features over direct training is particularly pronounced in low-data regimes, with sequential learning demonstrating the most consistent superiority across all sample sizes.}
\end{figure}

\subsection*{GenFAR features show strong generalization capability across studies compared to single-task CNNs}

We further assessed the generalization capability of sequential and independent GenFAR features across different sites compared to directly trained convolutional neural networks (CNNs).
This was performed by evaluating the generalization of the GenFAR feature representations across tasks as well as across studies and cohorts.
Performance was compared using a separate, external evaluation dataset from the Multi-Ethnic Study of Atherosclerosis (MESA, see Methods section) \cite{austin2022}.

The GenFAR features were learned on all other tasks (except the target task) from the primary training set. The direct training (5-fold cross-validation) performance of the secondary predictor was reported as a baseline. In the baseline experiments, a deep learning predictor was trained directly on the target task using data from the external study. Its 5-fold CV performance was compared to that of both the independent and sequential GenFAR-based approaches. Table 3 presents the results for different tasks, task types, and models. Both GenFAR approaches consistently outperform direct training, with the sequential framework demonstrating superior performance over the independent approach.

For the Age regression task, the sequential GenFAR model showed a substantial improvement in the correlation coefficient (0.701) compared to both the independent model (0.684) and the directly trained model (0.531). Similarly, in the BMI regression task, the sequential model achieved the highest correlation coefficient (0.598) compared to independent (0.582) and direct training (0.318).

For the classification tasks, the sequential framework similarly outperformed both alternatives. In the Hypertension classification task, the sequential model achieved an AUC of 0.738, compared to independent (AUC: 0.724) and direct training (AUC: 0.632). In the Smoking classification task, the sequential model attained an AUC of 0.584, versus independent (AUC: 0.571) and direct training (AUC: 0.529).

\singlespacing
\begin{center}
\resizebox{0.65\textwidth}{!}{%
\begin{tabular}{lcccccc}
\toprule
\multicolumn{7}{c}{\textbf{Classification Tasks}} \\
\midrule
Task & \multicolumn{3}{c}{AUC} & \multicolumn{3}{c}{Accuracy} \\
\cmidrule(lr){2-4} \cmidrule(lr){5-7}
& Direct & Independent & Sequential & Direct & Independent & Sequential \\
\midrule
Hypertension & 0.632 & 0.724 & 0.738 & 0.624 & 0.693 & 0.712 \\
Smoking & 0.529 & 0.571 & 0.584 & 0.526 & 0.568 & 0.581 \\
\midrule
\multicolumn{7}{c}{\textbf{Regression Tasks}} \\
\midrule
Task & \multicolumn{3}{c}{Correlation} & \multicolumn{3}{c}{MAE} \\
\cmidrule(lr){2-4} \cmidrule(lr){5-7}
& Direct & Independent & Sequential & Direct & Independent & Sequential \\
\midrule
Age & 0.531 & 0.684 & 0.701 & 5.631 & 5.212 & 5.102 \\
BMI & 0.318 & 0.582 & 0.598 & 3.890 & 3.126 & 3.021 \\
\bottomrule
\end{tabular}%
}
\end{center}
\vspace{12pt}
\nopagebreak
\noindent\small \textbf{Table 3: Performance comparison of sequential, independent, and direct training models in the external evaluation dataset.} A separate external evaluation dataset from the MESA study is used to evaluate the generalizability of GenFAR on various tasks (Age, BMI, Hypertension, and Smoking). Sequential and independent GenFAR models leverage generalized feature representations learned using data in the primary training set for all tasks except the target task, while the direct training model trains a CNN directly on the target task using external study data. Both GenFAR approaches consistently outperform direct training, with sequential learning demonstrating the strongest generalization capabilities across all evaluated tasks.

\doublespacing
\section*{Discussion}

Deep learning models have demonstrated significant potential in the field of medical imaging, primarily focusing on specialized single-task learning.
The emergence of foundational models within the broader deep learning community underscores the value of versatility and generalizability that large, multi-purpose systems can offer.
In this study, we introduced a deep learning-based neuroimaging model framework to extract a large and diverse set of clinically relevant representations from T1-weighted MRIs across a wide range of tasks.
Our approach contributes to the development of foundational representations of the brain by utilizing full-3D minimally preprocessed brain MRIs as input, allowing us to leverage the complete spatial context of the input MRI data.
We presented two complementary learning frameworks within GenFAR: an independent framework where tasks learn in parallel, and a sequential framework where tasks progressively build upon previously learned representations.
To ensure broad accessibility, we release the trained models through multiple channels: a public GitHub repository with self-contained inference pipelines for local deployment, and a cloud-based web portal that enables feature extraction without installation or dedicated computing resources.
These models are publicly available and applicable to standard structural T1 weighted MRI, making them immediately applicable in practical settings.

By leveraging an extensive and diverse dataset of 50,302 T1 brain MRIs across 12 studies, we demonstrated that features pertinent to specific neuroimaging tasks are often shared across other tasks and can be learned through them.
Our model integrates 17 clinically-relevant neuroimaging tasks, including demographic features (age, education years), clinical measurements (BNT, fluid intelligence, digit span forward, BMI, Total-Tau CSF, Abeta CSF, CDR, MMSE, smoking status, APOE genotype), and diagnoses (AD/MCI, diabetes, hyperlipidemia, hypertension, and depression), which were selected based on their diversity, data availability, and biological or clinical relevance.
The number of associated labeled samples varies between tasks ranging from 1,673 to 49,246 samples (\textbf{Table 4}), yet even at the lower end, the sample size remains large compared to similar works.
The combination of data from various sources has resulted in a dataset that covers diverse demographics, clinical measurements, and diagnostic categories.
This broad representation ensures that our model is exposed to a wide range of features, which facilitates learning and generalization across distinct tasks and populations.
Moreover, the dataset exhibits a range of acquisition-related factors, including multiple scanner manufacturers, models, field strengths (1.5 and 3 Tesla), and acquisition protocols.
This inherent variability in the data augments the robustness of our model, as it necessitates the identification of robust signal amidst such diversity.

The efficacy of deep learning predictors for medical imaging typically relies on a large training sample, especially when using full 3D scan volumes.
The scale of training data used in this work enabled the learning of powerful task-specific prediction channels under two frameworks.
The initial independent framework trained each task in parallel, establishing that the aggregate of task-specific learned representations could contribute valuable signal to other tasks.
However, this approach lacked direct knowledge sharing between the learning tasks.
To address this, we developed the sequential learning framework, designed to allow weaker tasks to bootstrap from the representations of stronger, preceding tasks.
Through extensive analysis of 5,000 random sequences, we determined that a sequence length of six was optimal, providing a balance between knowledge transfer and the risk of overfitting, which became pronounced in longer sequences (\textbf{Figure 2}).

To optimize the task ordering, we introduced the "Donor Score" metric, which quantified the marginal benefit of including one task in the sequence of another.
Certain tasks emerged as consistently strong positive donors, particularly Age, AD/MCI, MMSE, Hypertension, and Hyperlipidemia, providing beneficial features across diverse prediction targets (\textbf{Figure 3}).
Age in particular stood out as being a strong donor to all 16 other tasks.
This broad applicability reflects age's fundamental role in brain structure and function with the aging process encompassing widespread neuroanatomical changes that intersect with numerous pathological conditions \cite{liang2022,beck2022,cole2015,bashyam2020}.
The aging brain exhibits both global patterns (such as cortical thinning and ventricular enlargement) and focal changes (including hippocampal atrophy and white matter lesions) that serve as informative markers for diverse neurological and cognitive outcomes \cite{liang2022,habes2021}.
Conversely, certain tasks acted as negative donors, namely, Total-Tau CSF (likely due to limited training data introducing noise) and Smoking (potentially due to weak structural correlates on T1-MRI), both introducing detrimental variance rather than meaningful generalizable features.
This analysis informed our final model architecture, where we established a fixed base sequence using the five strongest donor tasks, with the remaining 10 non-excluded tasks (excluding Total-Tau CSF and Smoking) slotted into the sixth position to benefit from the strong foundation without introducing noise.

After deriving the prediction channels from both frameworks, we experimentally demonstrated the cross-task feature generalizability using a Leave-Task-Out (LTO) validation scheme.
These findings support the claim that the feature representations learned by GenFAR are generalizable and can serve as a foundation for secondary models in neuroimaging.
The results consistently showed that the sequential framework outperformed the independent framework, which in turn generally outperformed direct training on the 3D images (\textbf{Table 2}).
In 16 of the 17 tasks, the sequential learned features achieved performance that was comparable or superior to models trained directly on the imaging data.
Notably, in several classification tasks (Hypertension, Hyperlipidemia, Diabetes, and Smoking) we observed a substantial increase in performance.
This demonstrates that the structured knowledge transfer in the sequential model confers benefits beyond what is learnable from either the image data directly or from the parallel aggregation of independent features.

Additionally, we extensively evaluated the sample efficiency of our learned features compared to direct training.
We found that the use of GenFAR features resulted in reliably better performance as the training sample size for a new task was decreased.
In all four of the LTO experiments where direct training out-performed the learned feature representation (Abeta CSF, Age, BNT, and Total-Tau CSF), the learned feature representation eventually overcame this performance decrease as samples were limited.
For these four tasks, the threshold of training samples where the learned features out-performed direct training was between 200-2000 samples depending on the task (Figure 4).
The sample sizes, and especially the diversity of the data used in this study are not typical of predictive work in medical imaging.
The diagnosis-related tasks selected for this study have a large amount of labeled data, but for many prediction tasks, such quantity of data is not available or practical to obtain.
As a result, datasets of sizes up to a few hundred subjects are often used to train image-level deep learning predictive models in neuroimaging.
GenFAR features, particularly in the sequential configuration, showed a considerable improvement over direct training when the training sample size is small (<1000), making it a suitable option for those working with such a training sample size and for adapting pretrained models to new studies.
Even if the training set used in the secondary set is small, the GenFAR feature representation still benefits from the large dataset and diverse optimization targets used in its initial training.

The external validation on MESA data further substantiated the generalization capabilities of our approach, with sequential learning consistently outperforming both independent and direct training approaches (\textbf{Table 3}).
The substantial improvements in correlation coefficients on this entirely separate cohort demonstrated that the learned representations capture fundamental neurobiological patterns that generalize beyond study-specific variations in acquisition protocols and population characteristics.

While our model demonstrates promising results in the context of T1-weighted MRIs, several limitations warrant consideration.
The lack of densely labeled data prohibits direct learning of joint feature extractors that share low-level features.
Prior work has investigated approaches to this problem; however, solutions often sacrifice modularity and extensibility.
Additionally, the learned features, while highly predictive, are difficult to interpret in terms of specific neurobiological underpinnings.
Gradient-based saliency methods such as Grad-CAM \cite{chattopadhay2018gradcam} can highlight brain regions contributing to individual features (\textbf{Supplemental Figure S2}), but the representations remain fundamentally non-linear and distributed, characteristic of deep learning models, and do not map directly onto traditional neuroanatomical constructs.
Based on our proposed framework, we envision that as new datasets are made available to the field, additional, diverse tasks may be added to extend the descriptive dimensions of the GenFAR feature set.

The development of foundational models in neuroimaging presents numerous future directions.
Incorporating additional imaging modalities, such as T2-MRI, functional MRI, diffusion MRI, or PET scans, will further enhance predictive capabilities by providing complementary information about brain function and connectivity.
Additionally, data quality and preprocessing can have a significant impact on model performance.
Future work could explore the influence of various preprocessing pipelines and data augmentation techniques to optimize the model's performance and robustness.
Moreover, the current model architecture and training approach could be expanded to explore alternative approaches, such as 3D U-Net variants, vision-transformers, and self-supervised learning for potential improvements in generalizability and adaptability across different imaging tasks and domains.

Numerous clinical applications can arise from our model involving quantitative interpretation and prediction from brain MRI data.
By providing a versatile and generalizable framework for extracting clinically relevant features from neuroimaging data, our model can assist in early detection of pathological changes in the brain, leading to more timely interventions and improved patient outcomes.
Furthermore, the model's ability to integrate a wide range of neuroimaging tasks may facilitate the development of personalized treatment plans tailored to individual patients' needs and specific disease trajectories.
The availability of GenFAR through both local deployment and a cloud-based web portal ensures that these capabilities are accessible to research groups regardless of their computational infrastructure, potentially accelerating the translation of neuroimaging findings across diverse research settings.

\section*{Methods}

\subsection*{Dataset}

This study was trained and evaluated on a large and diverse dataset consisting of 50,302 T1 brain MRIs (49246 in the primary set and 1056 in a left-out evaluation set) with available concomitant label information for 17 clinically-relevant features from the iSTAGING consortium. These features included demographics such as age and years of education and clinical measurements such as smoking status, BNT, fluid intelligence, digit span forward, BMI, CSF Total-Tau, CSF Abeta, CDR®, MMSE, and APOE genotype. Additionally, the dataset included diagnosis information for AD/MCI, diabetes, hyperlipidemia, hypertension, and depression. The dataset comprised samples from 12 different neuroimaging studies (\textbf{Table 4}).

The availability of label data in our compiled dataset is contingent upon the information provided for our use by the original studies from which the MRIs were sourced and availability with our internal continuous data integration efforts, which may not be fully representative of the data available within each study. Each study has its unique set of objectives, patient cohorts, and data collection protocols, which in turn determines the type and granularity of the labels that are available. Consequently, the annotation richness and completeness can vary across studies, with some providing a more comprehensive set of labels than others. In some cases, the original studies might have focused on specific demographic features, clinical measurements, or diagnoses, leading to a scarcity of label data for certain tasks in our dataset. Conversely, other studies might have collected a wider range of labels, contributing to a more complete subset of tasks. The number of associated labeled samples varies widely between tasks, ranging from 1691 to 49246 samples.

The Wisconsin Registry for Alzheimer's Prevention (WRAP) \cite{johnson2018} is a longitudinal study examining Alzheimer's disease risk and progression through cognitive testing, imaging, genetics, biomarkers, and lifestyle factors.
The Biomarkers of Cognitive Decline Among Normal Individuals (BIOCARD) \cite{albert2014,gross2017} study is a longitudinal study of cognitively normal individuals with a family history of Alzheimer's disease, including cognitive testing, imaging, and biological samples to identify biomarkers of cognitive decline.
The Look AHEAD \cite{espeland2020} study was a randomized controlled trial examining the long-term effects of an intensive lifestyle intervention on cardiovascular outcomes in overweight and obese adults with type 2 diabetes.
The Coronary Artery Risk Development in Young Adults (CARDIA) \cite{cermakova2017} study is a multi-center longitudinal study examining cardiovascular disease risk factors and subclinical disease in a cohort of men and women followed since age 18-30.
The Australian Imaging, Biomarker, and Lifestyle (AIBL) \cite{pietrzak2015} study is a longitudinal study of aging and Alzheimer's disease including cognitive testing, imaging, and biomarkers in older adults with Alzheimer's disease, mild cognitive impairment, and healthy controls.
The Open Access Series of Imaging Studies (OASIS) \cite{marcus2007} is a compilation of imaging, biomarkers, and cognitive data on adults at various stages of cognitive decline collected over 30 years at Washington University.
The Baltimore Longitudinal Study of Aging (BLSA) \cite{ferrucci2008} is a longitudinal study of healthy aging led by the National Institute on Aging including imaging, biomarkers, genetics, and cognitive testing.
The PENN \cite{wolk2012} data is a clinical cohort of scans collected by the Penn Memory Center at the University of Pennsylvania with a focus on aging and Alzheimer's disease.
The Women's Health Initiative Memory Study (WHIMS) \cite{resnick2009} is an ancillary study to the Women's Health Initiative examining brain MRI scans in older women who participated in two randomized hormone therapy trials.
The Alzheimer's Disease Neuroimaging Initiative (ADNI) \cite{weiner2010,jack2008} is a public-private partnership study coordinated by the Alzheimer's Therapeutic Research Institute at the University of Southern California.
It focuses on collecting longitudinal imaging, biomarkers, genetics, and assessments in adults with Alzheimer's disease, mild cognitive impairment, and healthy controls.
UK Biobank \cite{alfaroalmagro2018} is a constantly updated large-scale biomedical database containing genetic, imaging, and health data from half a million UK participants funded by the UK Department of Health.

For regression tasks, we utilized all available labeled data, regardless of the distribution across different studies. However, for classification models, we ensured that every study had representation from both controls and patients, or the study was excluded. This was to avoid any scanner bias that might arise when the scanner variation could be mistaken as the classification signal. In the specific cases where a site solely included sparsely labeled data from the patient class, and the study's patient selection criteria assume that controls do not have that condition, we balanced the dataset by sampling an equal number of controls corresponding to the patient population from that particular site. This selection criterion was used to establish an unbiased learning and evaluation environment for the model while retaining as much labeled data as possible.

The dataset has been preprocessed according to a standardized pipeline, with procedures such as skull stripping, bias field correction, and affine registration to a template.

\singlespacing
{\setlength{\tabcolsep}{2pt}\renewcommand{\arraystretch}{1.0}\scriptsize
\begin{longtable}{C{1.6cm}C{1.0cm}C{1.0cm}C{1.0cm}C{1.0cm}C{1.0cm}C{1.0cm}C{1.0cm}C{1.0cm}C{1.0cm}C{1.0cm}C{1.0cm}C{1.1cm}C{1.0cm}}
\toprule
\scriptsize Task & \rotatebox{60}{\scriptsize WRAP~\cite{johnson2018}} & \rotatebox{60}{\scriptsize BIOCARD~\cite{albert2014,gross2017}} & \rotatebox{60}{\scriptsize lookAHEAD~\cite{espeland2020}} & \rotatebox{60}{\scriptsize CARDIA~\cite{cermakova2017}} & \rotatebox{60}{\scriptsize AIBL~\cite{pietrzak2015}} & \rotatebox{60}{\scriptsize OASIS~\cite{marcus2007}} & \rotatebox{60}{\scriptsize BLSA~\cite{ferrucci2008}} & \rotatebox{60}{\scriptsize PENN~\cite{wolk2012}} & \rotatebox{60}{\scriptsize WHIMS~\cite{resnick2009}} & \rotatebox{60}{\scriptsize ADNI~\cite{weiner2010,jack2008}} & \rotatebox{60}{\scriptsize UKBIOBANK~\cite{alfaroalmagro2018}} & \scriptsize \textbf{Total} & \rotatebox{60}{\scriptsize MESA~\cite{austin2022}} \\
\midrule
\endfirsthead
\toprule
\scriptsize Task & \rotatebox{60}{\scriptsize WRAP~\cite{johnson2018}} & \rotatebox{60}{\scriptsize BIOCARD~\cite{albert2014,gross2017}} & \rotatebox{60}{\scriptsize lookAHEAD~\cite{espeland2020}} & \rotatebox{60}{\scriptsize CARDIA~\cite{cermakova2017}} & \rotatebox{60}{\scriptsize AIBL~\cite{pietrzak2015}} & \rotatebox{60}{\scriptsize OASIS~\cite{marcus2007}} & \rotatebox{60}{\scriptsize BLSA~\cite{ferrucci2008}} & \rotatebox{60}{\scriptsize PENN~\cite{wolk2012}} & \rotatebox{60}{\scriptsize WHIMS~\cite{resnick2009}} & \rotatebox{60}{\scriptsize ADNI~\cite{weiner2010,jack2008}} & \rotatebox{60}{\scriptsize UKBIOBANK~\cite{alfaroalmagro2018}} & \scriptsize \textbf{Total} & \rotatebox{60}{\scriptsize MESA~\cite{austin2022}} \\
\midrule
\endhead
\bottomrule
\\[12pt]
\nopagebreak
\multicolumn{14}{p{0.95\textwidth}}{\small \textbf{Table 4: Distribution of labeled samples across neuroimaging tasks and studies used for training/validation.} Shown are the number of samples for each of the 17 neuroimaging tasks, broken down by the 11 core neuroimaging studies and the external evaluation set (MESA). The varying sample sizes reflect the differences in the original studies\textquotesingle{} objectives and data collection protocols, highlighting the diversity and heterogeneity of the compiled dataset. For regression tasks, we utilized all available labeled data, regardless of the distribution across different studies. However, for classification tasks, we ensured that every study had representation from both controls and patients or the study was excluded. This was to avoid any scanner bias that might arise when the scanner variation could be mistaken as the classification signal. In the specific cases where a site solely included sparsely labeled data from the patient class, and the study's patient selection criteria assume that controls do not have that condition, we balanced the dataset by sampling an equal number of controls corresponding to the patient population from that particular site. This selection criterion was used to establish an unbiased learning and evaluation environment for the model while retaining as much labeled data as possible. The data used in this study may not be fully representative of the totality of labeled data available within each study.} \\
\endlastfoot
\scriptsize MMSE & - & 300 & - & - & 929 & 1085 & 923 & 1182 & - & 1744 & - & \textbf{6163} & - \\
\scriptsize Education Years & - & 298 & - & 892 & 601 & 1086 & 1121 & 1221 & - & 1744 & - & \textbf{6963} & - \\
\scriptsize CDR Global & - & 301 & - & - & 739 & 1086 & 287 & 487 & - & 1774 & - & \textbf{4674} & - \\
\scriptsize Abeta CSF & - & 245 & - & - & - & - & - & 208 & - & 1238 & - & \textbf{1691} & - \\
\scriptsize Total-Tau CSF & - & 245 & - & - & - & - & - & 207 & - & 1221 & - & \textbf{1673} & - \\
\scriptsize Depression & - & - & - & 288 & 184 & - & - & 84 & 13 & 240 & 3697 & \textbf{4506} & - \\
\scriptsize BNT & - & 302 & - & - & 730 & 758 & 178 & 867 & - & 1730 & - & \textbf{4565} & - \\
\scriptsize Age & 272 & 309 & 311 & 892 & 975 & 1093 & 1129 & 1235 & 1418 & 2443 & 39169 & \textbf{49246} & 1056 \\
\scriptsize AD/MCI & - & 287 & - & - & 832 & 935 & 1114 & - & - & 2443 & - & \textbf{5611} & - \\
\scriptsize APOE Genotype & 267 & 306 & - & 658 & 767 & 1064 & 1008 & 339 & 1329 & 2164 & 33040 & \textbf{40942} & - \\
\scriptsize Diabetes & - & - & - & 155 & - & 202 & 26 & 45 & 134 & 240 & 3304 & \textbf{4106} & - \\
\scriptsize Hyperlipidemia & - & - & - & 860 & - & - & - & - & - & - & 16606 & \textbf{17466} & - \\
\scriptsize Hypertension & - & 6 & 309 & 776 & - & - & 1000 & - & - & 2404 & 18216 & \textbf{22711} & 1053 \\
\scriptsize BMI & 272 & 228 & 311 & 892 & 687 & 1013 & 1103 & 231 & 1418 & 2405 & 38280 & \textbf{46840} & 1056 \\
\scriptsize Smoking & 104 & 6 & 310 & 361 & 412 & 947 & 1060 & - & 663 & 1809 & 39330 & \textbf{45002} & 1055 \\
\scriptsize Digit Span Forward & 271 & 302 & - & - & 733 & 762 & 1104 & 493 & 97 & 819 & 26433 & \textbf{31014} & - \\
\scriptsize Fluid Intelligence & - & - & - & - & - & - & - & - & - & - & 36435 & \textbf{36435} & - \\
\bottomrule
\end{longtable}}
\doublespacing

\subsection*{External Evaluation Dataset}

The external evaluation dataset was used to assess the model\textquotesingle s generalizability. This dataset was from the MESA~\cite{austin2022} neuroimaging study and was not included in the main dataset used for training and validation. The primary aim of this evaluation was to determine the model\textquotesingle s performance when faced with previously unseen data from a different study, which may have unique acquisition protocols, imaging parameters, or patient populations. The external evaluation dataset was subject to the same preprocessing pipeline applied to the main dataset. Data availability allowed us to assess four tasks in the MESA sample: Age, hypertension, BMI, and smoking status.

MESA is an observational study of a multi-ethnic cohort of Americans free of clinically-recognized cardiovascular disease at enrollment in 2000-2002. The sample of the MESA~\cite{austin2022} dataset used was comprised of 1,056 participants with a mean age of 73.0 years (range 60-98 years) with neuroimaging in 2018-2019. The dataset included hypertension status, with 659 hypertensive participants and 394 non-hypertensive participants. BMI measurements ranged from 15.8 to 57, with a mean of 28.1. 500 participants reported never smoking, while 555 reported either former or current smoking.

\subsection*{Prediction Tasks}

The dataset utilized in our research encompassed an extensive array of tasks, carefully selected to balance data availability with inclusion of diverse aspects such as cognition, genetics, aging, disease, and biomarkers. These tasks encoded clinically significant factors that influence or are influenced by brain structure and function.

\textbf{Regression Task Variables:}

\begin{enumerate}
\item Boston Naming Test (BNT)~\cite{kaplan1983}: A measure of confrontation naming ability that assesses language comprehension and verbal expression. Participants identify visually presented objects ranging from common to rare items. Scored as the number of correctly named items (maximum 60).
\item Fluid Intelligence: Assessment of abstract reasoning and problem-solving ability independent of acquired knowledge. Measured using pattern recognition and logical reasoning tasks under time constraints. Higher scores indicate superior fluid reasoning capacity.
\item Digit Span Forward: A measure of short-term memory capacity and attention. Participants repeat sequences of digits presented at one-second intervals. Scored as the longest sequence correctly recalled.
\item Body Mass Index (BMI): An measure calculated as weight (kg) divided by height squared (m\textsuperscript{2}). Used to classify individuals as underweight ($<$18.5), normal weight (18.5-24.9), overweight (25-29.9), or obese ($\geq$30).
\item CSF Total-Tau (pg/ml): Cerebrospinal fluid concentration of tau proteins, a biomarker of neuronal injury and degeneration. Elevated levels indicate neuronal damage, particularly in neurodegenerative conditions.
\item CSF Amyloid-beta 42 (pg/ml): Cerebrospinal fluid concentration of amyloid-beta 42 peptide. Decreased levels indicate brain amyloid plaque accumulation, a hallmark of Alzheimer\textquotesingle s disease pathology.
\item Clinical Dementia Rating® (CDR®)~\cite{hughes1982} Global: A clinician-rated scale assessing dementia severity across six cognitive and functional domains (memory, orientation, judgment and problem-solving, community affairs, home and hobbies, and personal care). Scored from 0 (normal) through 0.5 (questionable impairment) to 1, 2, and 3 (mild, moderate, and severe dementia).
\item Years of Education: Total years of formal education completed.
\item Mini-Mental State Examination (MMSE)~\cite{folstein1975}: A cognitive measure assessing orientation, memory, attention, language, and visuospatial skills. Scored 0-30, with lower scores indicating greater impairment.
\item Age: Participant chronological age in years at time of imaging.
\end{enumerate}

\textbf{Classification Tasks Variables:}

\begin{enumerate}
\item Alzheimer\textquotesingle s Disease/Mild Cognitive Impairment (AD/MCI): Binary classification distinguishing individuals with AD or MCI from cognitively normal controls. AD represents progressive neurodegeneration with cognitive and functional decline. MCI indicates objective cognitive impairment without functional impairment sufficient for dementia diagnosis.
\item APOE Genotype: Binary classification of apolipoprotein E $\varepsilon$4 allele carrier status. APOE has three common alleles ($\varepsilon$2, $\varepsilon$3, and $\varepsilon$4). Classification distinguished between no $\varepsilon$4 alleles present versus 1 or 2 $\varepsilon$4 alleles present. The $\varepsilon$4 variant confers increased risk and earlier onset of Alzheimer\textquotesingle s disease.
\item Diabetes: Binary classification of diabetes, a metabolic disorder characterized by chronic hyperglycemia. Diabetes labels for BLSA and ADNI were inferred based on fasting blood glucose measures and/or use of glucose-lowering agents. All other cohorts relied on participant self-report of professional diagnosis. All Look AHEAD participants were diabetic, so the study was excluded due to absence of non-diabetic controls.
\item Hyperlipidemia: Binary classification of abnormally elevated blood lipid levels, including cholesterol and triglycerides, representing increased cardiovascular disease risk.
\item Hypertension: Binary classification of chronically elevated blood pressure. Hypertension labels for CARDIA and BLSA were inferred based on antihypertensive medication use, systolic blood pressure $>$160 mmHg, or diastolic blood pressure $>$95 mmHg. Other cohorts relied on clinical diagnosis or self-report.
\item Depression: Binary classification based on participant-reported medical history of major depressive disorder, characterized by persistent mood disturbance, anhedonia, and associated cognitive and somatic symptoms.
\item Smoking: Binary classification of tobacco use history, distinguishing never-smokers from current or former smokers.
\end{enumerate}

\subsection*{Sampling Strategy}

To address the significant class sample size imbalance in many of the available classification tasks, we employed a balanced sampling approach to create the training set. This method ensured that each class was equally represented in the training set, mitigating the influence of majority classes on the model\textquotesingle s performance. The number of samples in the smallest class served as a baseline for the number of samples to include from each class. We retained all samples from the smaller class and randomly selected an equal number of samples from the larger class.

For regression tasks, we utilized the entire available dataset without any additional sampling. The model was trained using all available samples, allowing it to learn the underlying relationships between input features and target variables.

\subsection*{Network Architecture}

GenFAR employed both independent and sequential frameworks to learn task-specific feature representations from 3D T1-weighted brain MRIs. In the independent framework, each of the 17 prediction channels consists of a 3D Squeeze-and-Excitation (SE) ResNet~\cite{hu2018squeeze} with a fully connected layer as the prediction head for the given task. The architecture follows a ResNet-18 style design adapted for volumetric data, accepting single-channel 3D MRI volumes (128$\times$128$\times$128, 1$\times$1$\times$1 mm\textsuperscript{3} isotropic voxels) as input.

The initial stem block consists of a 7$\times$7$\times$7 3D convolution with 32 output channels and stride 2, followed by 3D batch normalization, LeakyReLU activation, and 3$\times$3$\times$3 max pooling with stride 2. This stem block reduces the spatial dimensions by a factor of 4 while establishing initial feature representations. The network contains four sequential residual stages with [2, 2, 2, 2] blocks per stage, progressively increasing channel dimensions from 32 to 64 to 128 to 256. Each residual block comprises: (1) a 3$\times$3$\times$3 convolution followed by batch normalization and LeakyReLU activation, (2) a second 3$\times$3$\times$3 convolution followed by batch normalization, (3) a Squeeze-and-Excitation attention module that adaptively recalibrates channel-wise feature responses, and (4) a residual connection with LeakyReLU activation. The network concludes with adaptive average pooling and produces a 512-dimensional (or 64-dimensional for the 6th task in sequential learning) output vector, which is input to the fully connected prediction head. The fully connected layer output uses a linear activation function for regression tasks and a sigmoid activation function for classification tasks. After training, the prediction heads are removed and the convolutional weights are retained for feature extraction.

In the sequential framework, the architecture mirrors the independent setup with key differences to enable progressive transfer of content from earlier tasks. The first task in a sequence operates identically to the independent framework, receiving only the 3D brain scan and producing an n-dimensional feature representation (n is 512 for the first 5 tasks and 64 for the 6th task). For subsequent tasks (e.g., task 2), the network receives both the 3D brain scan and the feature representations from all preceding tasks. The feature representations from prior tasks are concatenated to the output of the 3D SE-ResNet before the fully connected layers to allow bootstrapping from prior learnings. This process iteratively builds richer representations, with later tasks learning only the incrementally beneficial features beyond those already provided. To determine optimal configurations, we analyzed 5,000 random sequences (see Training section below). The final sequential model uses a fixed base sequence of the 5 strongest donor tasks (Age, AD/MCI, MMSE, Hypertension, Hyperlipidemia; see Donor Score section below). Each remaining non-strictly negative task is slotted into the 6th position of this base sequence to generate its feature representation (dimension 64), benefiting from the base without introducing noise or overfitting into the chain. Tasks with strongly negative Donor Scores (Total-Tau CSF and Smoking) are excluded. This produces feature representations for the 5 base tasks plus 10 slotted tasks (e.g., base representations plus those from slotted tasks), concatenated to form the GenFAR features (totaling 3,200 features from 5 sets of 512 features and 10 sets of 64 features).

After each prediction channel is optimized on its respective tasks (in either framework), the weights from the prediction channels are frozen to serve as the foundation for the secondary prediction layers. The outputs from all the prediction channels are concatenated before being input into a new fully connected encoder. This new encoder can then be trained on a new arbitrary task. The final GenFAR network consists of 15 or 17 parallel prediction channels and a one or two-layer fully connected encoder.

\subsection*{Sequence length selection}
To determine optimal configurations for sequential learning, we conducted extensive experiments with 5,000 randomly generated task sequences. Initial attempts at learning-based approaches, including Thompson sampling and evolutionary algorithms, resulted in severe overfitting to the validation set. The repeated evaluation of sequences during optimization on the validation set results in overfitting to it, leading to poor test generalization.

Instead, we trained sequential models on 5,000 task sequences of lengths 3-9 (with smaller samples for lengths 12 and 16 to assess effects), using random orderings without optimization. Aggregate analysis of these sequences identified an optimal task length of 6. We compared model performance across all lengths using pairwise comparisons (Wilcoxon signed-rank tests) to statistically verify if the best-performing length (Length 6) was significantly better than the others, correcting for multiple tests to ensure robustness. This length showed statistically significant superior performance (p < 0.05 after FDR correction) compared to other lengths, balancing knowledge transfer benefits against noise accumulation and overfitting (see Figure 2 in Results). Longer sequences (e.g., 9+) exhibited high validation performance but poor test generalization.

\subsection*{Donor Score}
To quantify task ordering and influence in the sequential framework, we developed the Donor Score metric (see Figure 3A in Results). The Donor Score quantifies the marginal benefit a task provides when included in a learning sequence.

Let \(P(r|s)\) denote the performance improvement (\%) of receiver task \(r\) when trained using sequence \(s\), relative to direct training baseline. For a given donor-receiver pair, we define:

\[\text{DonorScore}(d, r) = \text{avg}[P(r|s) : d \in s] - \text{avg}[P(r|s) : d \notin s]\]

where the first term averages performance across all sequences containing donor task \(d\), and the second term averages across sequences without \(d\). A positive Donor Score indicates that task \(d\) contributes more to task \(r\)'s performance than the average task, suggesting beneficial knowledge transfer. Negative scores indicate that including task \(d\) is detrimental compared to other tasks.

To evaluate statistical significance of donor effects, we conducted independent samples t-tests (Welch's t-test) comparing sequences containing versus not containing each donor task. This test was chosen as it does not assume equal variances between groups. The analysis included 272 unique donor-receiver pairs across 17 tasks, with sequences limited to length \(\leq\)9 (based on the optimal sequence length findings presented in Figure 2). Given the large number of simultaneous tests, we applied FDR correction for multiple comparisons to control the expected proportion of false positives while maintaining statistical power. All statistical analyses were performed using Python 3.9 with scipy.stats for t-tests and statsmodels for multiple comparison corrections.

Based on the Donor Score analysis (Figure 3B-D), we established a fixed base sequence using the five strongest positive donor tasks: Age (mean Donor Score: 0.12), AD/MCI (0.10), MMSE (0.09), Hypertension (0.07), and Hyperlipidemia (0.06). These tasks were selected based on their consistent positive contributions across multiple receiver tasks and rare negative effects (Figure 3C-D). Conversely, tasks with strongly negative Donor Scores (Total-Tau CSF and Smoking) were excluded. Each of the remaining 10 tasks was then inserted as the sixth task in this sequence, allowing it to benefit from the strong base features while preventing harmful tasks from disrupting the feature hierarchy. This configuration balances knowledge transfer benefits against noise accumulation and overfitting risks identified in our sequence length experiments and forms the basis for the sequential learning results.

\subsection*{Training}
Each task-specific network is individually optimized for its prediction endpoint using the Adam optimizer~\cite{kingma2014}. Classification endpoints use binary cross entropy loss and regression endpoints use mean squared error loss. Data for each task is randomly split into training (65\%), validation (15\%), and test (20\%) sets. The training set are the samples whose loss is directly optimized on and to which the weights of the network are fit. The loss and other metrics for the validation set are evaluated at the end of each epoch, measuring how well the weights learned on the training set generalize to unseen samples. The convergence criteria and optimal model state is determined on the validation set. A network is considered to be converged when after 5 consecutive epochs there is no improvement in the validation loss. The best performing epoch (based on a pre-defined selection metric evaluated on the validation set) for the network is then chosen as the final model. For classification tasks this metric is AUC and for regression tasks this metric is the Pearson correlation coefficient. After the best epoch is chosen, the selected model is then evaluated on the held-out test set to calculate the final reported performance of the network.

The internal feature representation learned can be used in two ways. For a new data sample, the scan is run though each of the prediction channels resulting in 17 or 15 sets of features (8,704 or 3200 features, depending on the independent or sequential framework). These features can then be used for analysis or in traditional ML techniques. In addition to feature extraction, the weights in the learned prediction channels can be frozen, and a new fully connected prediction head that inputs the aforementioned features can be added. The topology of this fully connected head can vary depending on the needs of the user. If the user wants to perform a regression or classification tasks they can use an output layer of dimension 1 (regression or binary classification) or larger (multi-class classification). 

\subsection*{Leave-Task-Out Cross-Validation}

We performed leave-task-out cross-validation (LTO-CV) to evaluate the generalizability of the learned features from the prediction channels in both independent and sequential frameworks. In this procedure, a single task, along with its corresponding evaluation data, is held-out while all other tasks are trained. This process results in 16 prediction channels, none of which have explicit training towards the held-out task or exposure to the held-out data. The prediction heads of these individual channels are removed, and their weights are frozen. A fully connected encoder is added after the concatenated output of these channels. The non-frozen encoder head is then optimized for the held-out prediction task using the corresponding held-out subjects. The performance difference between the original direct trained model (with matched training subjects) and the model with a particular task held out during prediction channel training is used to determine the extent to which features learned in other tasks generalize to the new task. This procedure is repeated 16 more times until each task has been evaluated as the left-out task.

For the independent framework, all 16 non-held-out tasks are trained in parallel as described in the Network Architecture and Training sections, producing 16 sets of 512-dimensional features that are concatenated and input to the secondary encoder for prediction on the held-out task.

For the sequential framework, the procedure adapts the fixed base sequence to maintain knowledge transfer while excluding the held-out task. If the held-out task is not one of the 5 base donor tasks (Age, AD/MCI, MMSE, Hypertension, Hyperlipidemia), the full base sequence of 5 tasks is trained, and each of the remaining 9 non-held-out, non-base tasks is slotted into the 6th position to generate its 64-dimensional feature representation. If the held-out task is one of the 5 base donor tasks, the base sequence is adjusted to the remaining 4 base tasks, each of the 10 other non-base task is slotted into the 5th position (resulting in a sequence length of 5). Note that the Total-Tau CSF and Smoking tasks are excluded from the features representations however they are included in the LTO evaluation to provide a comprehensive assessment.

The data split between the training and held-out tasks is crucial to prevent data leakage and ensure sample availability balance. Since individual samples often serve as training data in multiple tasks in the full model, when they have multiple associated data labels/values, prioritizing the balance between sample availability in the training tasks and the held-out task is necessary. If a sample is to be used in the held-out set, it cannot be used in the training of any of the directly trained prediction channels. For a particular held-out task, we reserve all training samples, up to a limit of 8,000 total samples, for the held-out set. This approach prevents tasks with widely available labels from overly limiting the data size of the directly trained prediction channels. For instance, in the case of age regression, age is available for all scans; therefore, the total number of held-out samples is limited to 8,000 (randomly selected). All remaining samples are used to train the in-sample prediction channels.

Given this split strategy, certain tasks with highly overlapping label availability may not contain enough data to be used in the training phase after task-specific data has been held out. In such cases, the in-sample task is dropped from the training if there are less than 300 samples left. For example, Abeta CSF and Tau CSF have highly overlapping label availability (and well under 8,000 samples), so when either is held out, the other is not included in the directly trained tasks. Training data that remains, after curation to eliminate data leakage, is split into the training (75\%) and validation (25\%) sets for each prediction channel. For the held-out tasks, 5-fold cross-validation is performed to determine the final performance of the task.

\subsection*{Technical Details / Computational Requirements}

Models were trained using 16-bit mixed precision to optimize both training speed and memory usage, as full brain 3D inputs demand substantial GPU memory. The training configuration included an initial learning rate of 0.0006, cosine annealing, and a batch size of 4. Stochastic weight averaging was incorporated after observing positive results in preliminary tests, contributing to improved generalization performance. Our deep learning framework was Pytorch~\cite{paszke2019}, and we employed MONAI~\cite{cardoso2022} for data augmentation purposes. For organization, reproducibility, and training speed, we integrated components of Pytorch Lightning~\cite{falcon2019}. MLOps and experiment tracking were facilitated by Weights \& Biases. Each model was trained on a single GPU (16GB or 40GB VRAM) with 64GB of allocated RAM and 8 allocated CPU cores. We utilized 8 threads of asynchronous data loading for fetching and transforming data, resulting in enhanced GPU utilization.

This work was carried out on the CUBIC High Performance Computing Cluster, which is equipped with 100+ GPUs. The training time for each individual prediction channel (using a single GPU) varied from 3 hours to 48 hours, depending on the data availability for a specific task.

\subsection*{Public Release and Accessibility}

To maximize the utility of GenFAR for the research community, we provide multiple avenues for accessing the trained models and extracting features from new data. The final model weights, along with self-contained inference pipelines for both GPU and CPU environments, are publicly released on GitHub (https://github.com/vishnubashyam/GenFAR\_Main) under an Open Responsible AI License (OpenRAIL). This allows researchers to deploy GenFAR locally within their own computational infrastructure.

To further lower barriers to adoption, we additionally provide a cloud-based web portal via NiChart (\url{https://neuroimagingchart.com/portal}) that enables feature extraction without local installation or dedicated computing resources. The web portal, hosted on Amazon Web Services, allows users to submit raw T1-weighted images through a drag-and-drop interface and receive the extracted GenFAR feature set. This approach ensures that GenFAR is accessible to researchers regardless of their computational resources or technical expertise.
  
\section*{Data Availability}

All data was obtained in a deidentified form from the respective study data repositories. Each component study received approval from appropriate review and regulatory bodies and ensured compliance with ethical standards and informed consent. Requests for data from each individual study may be made directly to the study oversight committee.

\section*{Code Availability}

The code used in this manuscript as well as a self-contained GPU and CPU inference pipeline is available on GitHub (\url{https://github.com/vishnubashyam/GenFAR_Main}) and released under an Open Responsible AI License (OpenRAIL).

\section*{Acknowledgments}

This research is associated with the following grant numbers:

NIH - RF1AG054409, U24NS130411, R01AG059869, U01AG068057, HHSN-260-2004-00012C, P30 AG066444, P01 AG026276, P01 AG03991, P50 AG05681, P01 AG03991, R01 AG021910, P50 MH071616, U24 RR021382, R01 MH56584, P30AG534255;

NIA, NSF - 191026, 206795

MESA is supported by contracts 75N92025D00022, 75N92020D00001, HHSN268201500003I, N01-HC-95159, 75N92025D00026, 75N92020D00005, N01-HC-95160, 75N92020D00002, N01-HC-95161, 75N92025D00024, 75N92020D00003, N01-HC-95162,  75N92025D00027, 75N92020D00006, N01-HC-95163, 75N92025D00025, 75N92020D00004, N01-HC-95164, 75N92025D00028, 75N92020D00007, N01-HC-95165, N01-HC-95166, N01-HC-95167, N01-HC-95168 and N01-HC-95169 and grant R01HL127659 from the National Heart, Lung, and Blood Institute, and by grants UL1-TR-000040, UL1-TR-001079, and UL1-TR-001420 from the National Center for Advancing Translational Sciences (NCATS).

V.B. is supported by the Ruth L. Kirschstein National Research Award (NIH T32-EB020087, PD: Felix W. Wehrli)

This research has been conducted using the UK Biobank Resource under application number 35148.

Data used in the preparation of this article was obtained from the Australian Imaging Biomarkers and Lifestyle flagship study of ageing (AIBL) funded by the Commonwealth Scientific and Industrial Research Organisation (CSIRO) which was made available at the ADNI database (\url{www.loni.usc.edu/ADNI}). The AIBL researchers contributed data but did not participate in analysis or writing of this report. AIBL researchers are listed at \url{www.aibl.csiro.au}.

Data used in preparation of this article were obtained from the Alzheimer's Disease Neuroimaging Initiative (ADNI) database (adni.loni.usc.edu). As such, the investigators within the ADNI contributed to the design and implementation of ADNI and/or provided data but did not participate in analysis or writing of this report. A complete listing of ADNI investigators can be found at: \url{http://adni.loni.usc.edu/wp-content/uploads/how_to_apply/ADNI_Acknowledgement_List.pdf}

The authors would like to acknowledge the clinical and neuropathology diagnostic support provided by the Wisconsin ADRC's Clinical, Neuropathology and Biomarkers Cores, and biostatistical support provided by the Data Management and Biostatistics Core.

The opinions and conclusions contained in this publication are solely those of the authors, and are not necessarily endorsed by the associated studies, institutions, and funding agencies and should not be assumed to reflect their opinions or conclusions.

\section*{Competing Interests}

The authors declare that they have no conflict of interest.

% References and supplemental materials (single-spaced)
\singlespacing
\bibliographystyle{unsrt}
\bibliography{bib/references}

\clearpage

\section*{Supplemental Materials}

\noindent\textbf{Supplemental Text 1: Task-specific prediction channels learn representations that probe the brain across various regions, contributing to predictive performance on new tasks}

We analyzed the relationship between learned features for a particular task and the volume of predefined neuroanatomical regions of interest (ROIs) to identify which brain regions are associated with the learned features (\textbf{Supplemental Figure S1A}). Note that this analysis only applies to the independent prediction channels. By looking at the maximum correlation coefficient between the various features learned in a prediction channel and the volumes of ROIs, across different tasks we see that the features learned in different tasks probe the brain across various regions. We note that this analysis is limited in that it only reflects the association of regional volumetric changes with learned features and does not capture all of the information that the learned features may contain. The learned features may also encode interactions between various regions of the brain, texture and shape information, and other types of information that cannot be captured by looking at the volume of individual ROIs in isolation. It is also worth noting that, when using the full 3D brain volume as input, it is likely that the prediction channels will learn a wide variety of features beyond just those that are associated with individual ROIs.

In \textbf{Supplemental Figure S1B,} we see the normalized predictive performance of each prediction channel for each target task. We see that many tasks learn features that are useful in unrelated tasks, pointing to the usefulness of cross-task knowledge transfer. Supplemental Figure S1B shows that some tasks, e.g. age prediction, produce feature sets that are useful across a variety of other tasks (the first row has many dark squares), whereas other tasks produce feature sets of less general value (e.g. the depression task row has relatively few dark squares). Since various tasks have different degrees of separability based on T1-MRI data, we normalize across the features learned for each target task, showing their comparative predictive performance. The contributions of each of these prediction channels, jointly, encode the predictive performance seen in \textbf{Table 1}.

\clearpage

\noindent\textbf{Supplemental Figure S1.} \textbf{(A) Influence of regional tissue volumes on each prediction channel}. The correlation between the learned features for a particular task and the volume associated with predefined neuroanatomical regions of interest (ROIs), shows the brain regions that are associated with learned features. Note that this analysis only uses features from the independent prediction channels as the sequential framework limit the interpretability of the features. We note that this analysis only associates regional volumetric changes with learned features and does not represent the full descriptive capacity of the learned features. Given the full 3D brain volume as input, we expect prediction channels to learn a large variety of other features, including interactions between various regions of the brain, texture and shape information, etc. The maximum correlation across learned features for each brain region is shown. \textbf{(B) Normalized relative prediction performance for features of each prediction channel to each target task.} Note that this analysis only applies to the independent prediction channels. The y-axis shows the prediction channel from which the features are learned. The x-axis shows the target task we are predicting using the learned features. The prediction performance is normalized across each column to show the comparative predictive performance of features learned in each prediction channel to the target task, while controlling for the variability in task difficulty. Red squares indicate when insufficient subjects were available due to overlapping samples and limited labeled data.

\begin{center}
\includegraphics[width=0.6\textwidth]{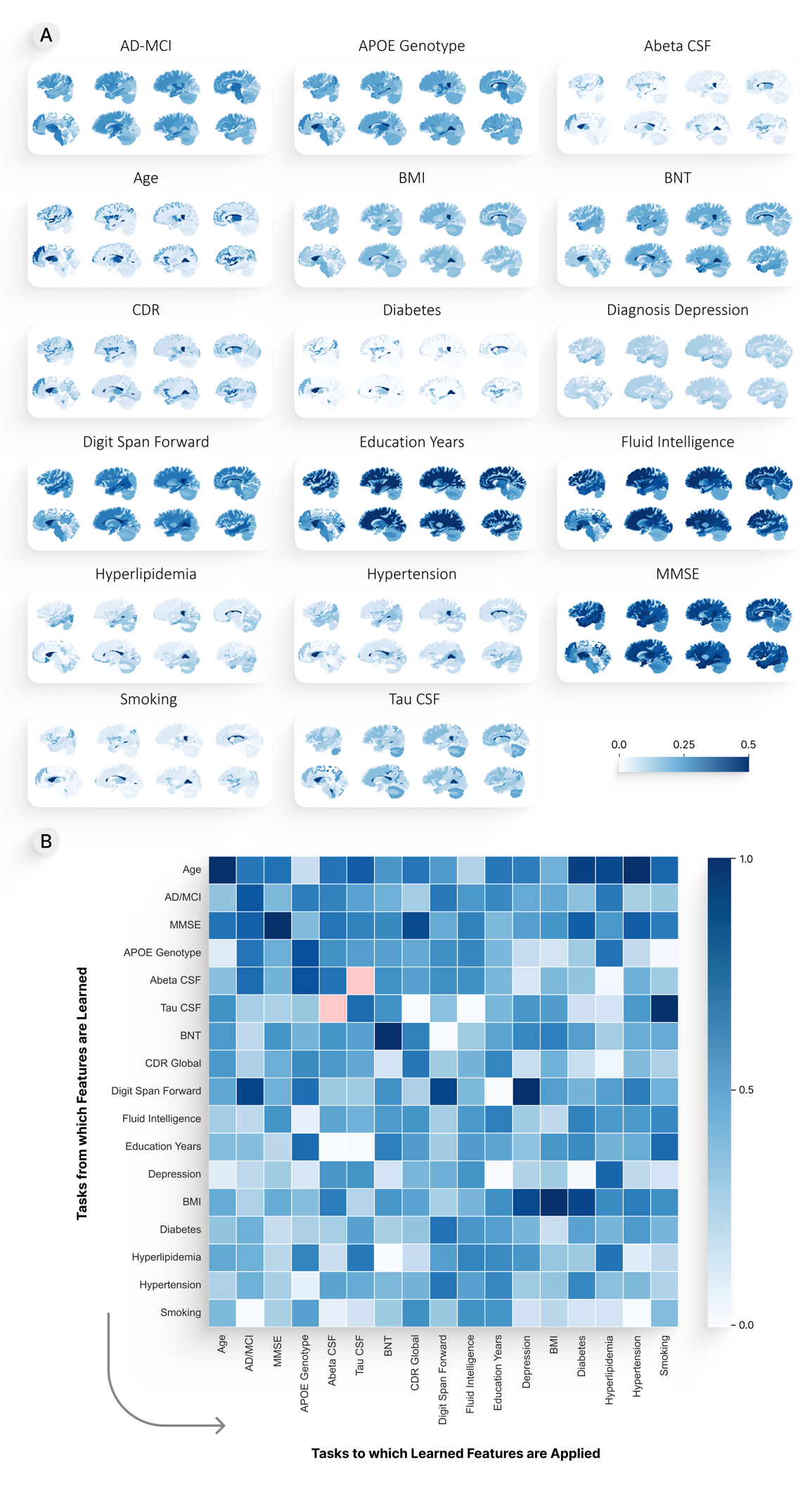}
\end{center}

\clearpage

\noindent\textbf{Supplemental Figure S2. Age-stratified Grad-CAM saliency maps for a representative learned feature.} Gradient-weighted Class Activation Mapping (Grad-CAM) visualization for a randomly selected feature from the age prediction channel, stratified by age group (Low, Medium, High). Each panel shows axial slices at eight standardized positions through the brain volume (slices 20, 35, 51, 67, 82, 98, 114, 130). Warmer colors (yellow) indicate regions with higher saliency contributing to the feature activation, while cooler colors (purple) indicate lower saliency. While such visualizations provide some insight into which brain regions contribute to learned features, they illustrate the challenge of interpreting deep learning representations: the features capture complex, distributed, and highly non-linear patterns that do not map directly onto traditional neuroanatomical parcellations or simple biological constructs.

\begin{center}
\includegraphics[width=0.85\textwidth]{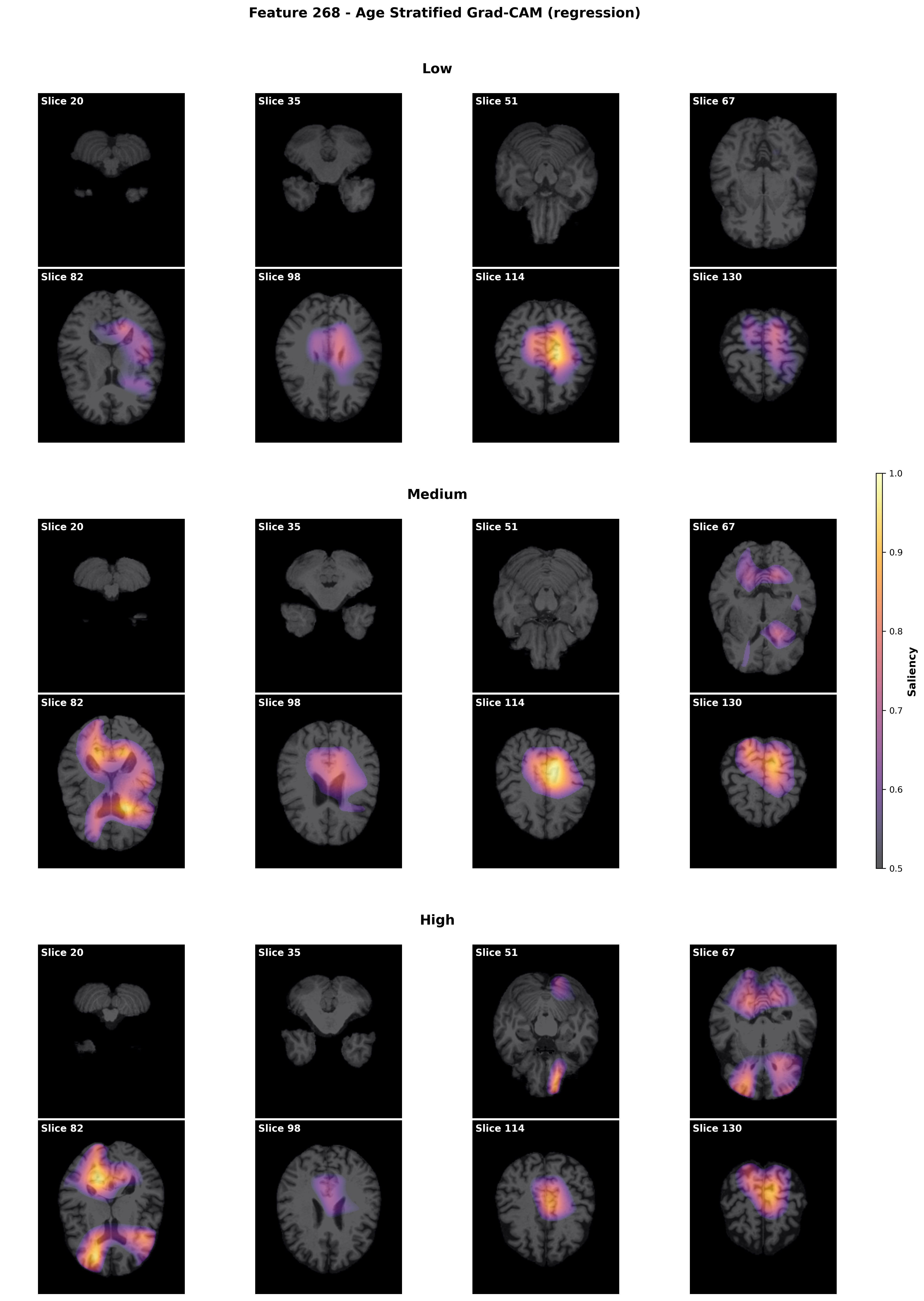}
\end{center}

\clearpage

\clearpage

\end{document}